\documentclass{article}

\usepackage{amsmath,amsfonts,bm}

\def\eqref#1{equation~\ref{#1}}

\def\1{\bm{1}}

\def\rmK{{\mathbf{K}}}

\def\rmQ{{\mathbf{Q}}}

\def\rmS{{\mathbf{S}}}

\def\rmV{{\mathbf{V}}}
\def\rmW{{\mathbf{W}}}

\def\va{{\bm{a}}}
\def\vb{{\bm{b}}}

\def\vg{{\bm{g}}}

\def\vx{{\bm{x}}}

\def\vz{{\bm{z}}}

\DeclareMathAlphabet{\mathsfit}{\encodingdefault}{\sfdefault}{m}{sl}
\SetMathAlphabet{\mathsfit}{bold}{\encodingdefault}{\sfdefault}{bx}{n}

\def\gD{{\mathcal{D}}}

\def\gL{{\mathcal{L}}}

\def\sR{{\mathbb{R}}}

\newcommand{\E}{\mathbb{E}}

\usepackage{arxiv}

\usepackage[utf8]{inputenc} 
\usepackage[T1]{fontenc}    
\usepackage{hyperref}       
\usepackage{url}            
\usepackage{booktabs}       
\usepackage{amsfonts}       
\usepackage{nicefrac}       
\usepackage{microtype}      
\usepackage{lipsum}
\usepackage{graphicx}
\graphicspath{ {./images/} }
\usepackage[utf8]{inputenc} 
\usepackage[T1]{fontenc}    
\usepackage{hyperref}       
\usepackage{url}            
\usepackage{booktabs}       
\usepackage{amsfonts}       
\usepackage{nicefrac}       
\usepackage{microtype}      
\usepackage{xcolor}         
\usepackage{wrapfig}

\usepackage{floatrow}
\newfloatcommand{capbtabbox}{table}[][\FBwidth]
\usepackage{wrapfig}
\usepackage{microtype}
\usepackage{url}
\usepackage{graphicx}
\usepackage{subfigure}
\usepackage{booktabs} 
\usepackage{multirow}
\usepackage{multirow}
\usepackage{float}
\usepackage{algorithm}
\usepackage{algpseudocode}
\usepackage{multicol}
\usepackage{makecell}
\usepackage{arydshln}
\usepackage{booktabs} 
\usepackage{amssymb}
\usepackage{pifont}
\usepackage{hyperref}

 \usepackage{natbib}

\title{Gradient Inversion Transcript: Leveraging Robust Generative Priors to Reconstruct Training Data from Gradient Leakage}

\author{
 Xinping Chen \\
  Department of Computer Science\\
  City University of Hong Kong\\
  83 Tat Chee Avenue, Kowloon Tong, Hong Kong\\
  \texttt{xinpichen2-c@my.cityu.edu.hk} \\
   \And
 Chen Liu \\
  Department of Computer Science\\
  City University of Hong Kong\\
  83 Tat Chee Avenue, Kowloon Tong, Hong Kong\\
  \texttt{chen.liu@cityu.edu.hk} \\
}

\begin{document}
\maketitle
\begin{abstract}
  We propose Gradient Inversion Transcript (GIT), a novel generative approach for reconstructing training data from leaked gradients. GIT employs a generative attack model, whose architecture is tailored to align with the structure of the leaked model based on theoretical analysis. Once trained offline, GIT can be deployed efficiently and only relies on the leaked gradients to reconstruct the input data, rendering it applicable under various distributed learning environments.
  When used as a prior for other iterative optimization-based methods, GIT not only accelerates convergence but also enhances the overall reconstruction quality. GIT consistently outperforms existing methods across multiple datasets and demonstrates strong robustness under challenging conditions, including inaccurate gradients, data distribution shifts and discrepancies in model parameters.
\end{abstract}


\section{Introduction} \label{sec:intro}

In distributed learning, each client trains its model on local data and shares the gradients with a central server, which aggregates them to update the global model \citep{jochems2016distributed, mcmahan2017communication, yang2019ffd}. 
Gradient sharing is also common in federated learning (FL)~\citep{NEURIPS2021_3b3fff64}, but unlike distributed learning, which involves a more centrally coordinated distribution of data across clients, federated learning focuses on preserving data privacy by ensuring that client data remains localized.
While these methods are effective in improving performance and efficiency without directly exposing the client's data to public, recent research~\citep{phong2017privacy, NEURIPS2019_60a6c400, zhao2020idlg} has shown that the gradients leaked by sharing can still cause sensitive information leakage, as attackers may exploit them to reconstruct the original training data used by the individual client, posing significant privacy risks in real-world learning systems.

There is a considerable amount of works proposed to reconstruct the training data from its gradient~\citep{NEURIPS2021_3b3fff64, phong2017privacy, NEURIPS2019_60a6c400, zhao2020idlg, wei2020framework, geiping2020inverting, zhu2020r, wang2020sapag, yin2021see, wang2023reconstructing, jeon2021gradient, li2022auditing, fang2023gifd, wu2023learning, chen2024recovering, wu2025dggi} based on varying levels of model access, as shown in Table~\ref{tab:my_label}.
These works generally fall into two major categories: iterative optimization methods, which iteratively optimize the reconstructed data to align its gradients with the leaked ones; and generative methods, which leverage a generative model to approximate the user data. 
Generative methods can be further subdivided into two types: latent space-based and generative model-based. The first type trains a latent space as the input of a pre-trained generative model. The second type trains a generative model to map leaked gradients to the corresponding user data.

Iterative optimization methods typically require repeated access to gradients from the target model~\citep{phong2017privacy, NEURIPS2019_60a6c400, wei2020framework, geiping2020inverting, wang2020sapag, NEURIPS2021_3b3fff64, chen2024recovering} or full access to the model parameters~\citep{zhu2020r, wang2023reconstructing}.
In contrast, the latent space-based methods~\citep{yin2021see, jeon2021gradient, li2022auditing, fang2023gifd, wu2025dggi} relies on a pre-trained generative model and public data that closely matches the distribution of the user data to train the latent space.
The generative model-based methods~\citep{wu2023learning} require input-gradient pairs from public datasets to train the generative model.

We focus on the generative models for input data reconstruction in this work. We call the model under attack the \textit{leaked model}.
The existing generative model-based method, such as LTI~\citep{wu2023learning}, usually employs a fixed-architecture multi-layer perception (MLP)~\citep{rosenblatt1958perceptron} as the generative model. However, this approach lacks justification for how gradients relate to the input data. We argue that the generative model to reconstruct the input data from the gradient should approximate the inverse of the gradient computation process and thus be adaptive to the architecture of the leaked model.
In this context, we introduce \textbf{gradient inversion transcript (GIT)} in this work to adaptively choose the architecture of the generative model to improve its effectiveness.
In addition, we can combine GIT with iterative optimization methods like IG~\cite{geiping2020inverting}, in which we use GIT's output as the initial estimation of the input data for iterative optimization methods to further refine the reconstructed input data.
                                                
\begin{figure}
    \centering\includegraphics[width=0.65\linewidth]{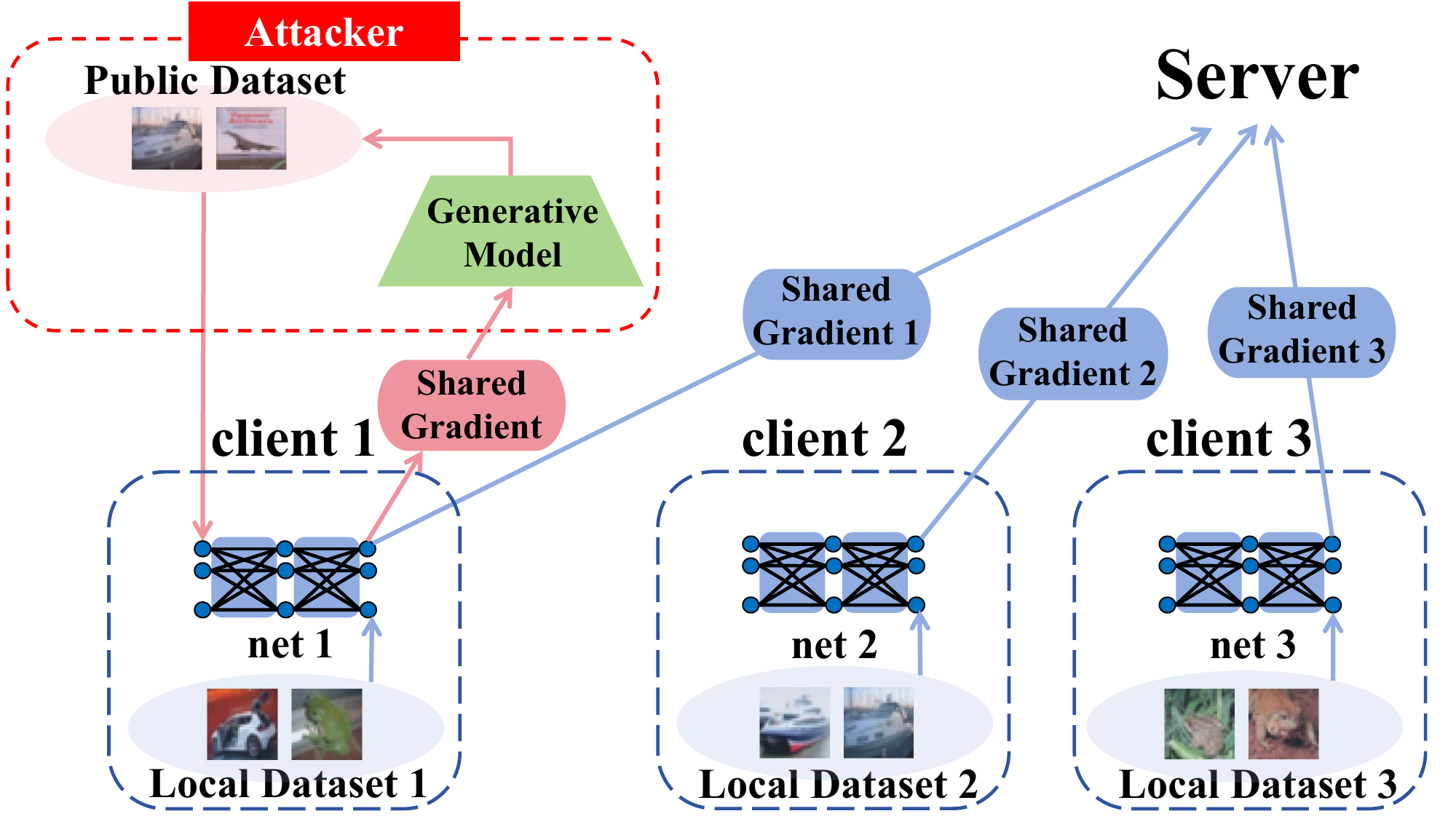}
    \caption{A flowchart of problem settings for GIT. The attacker hacks the channel of one client to inject data and utilizes the obtained input-gradient pair to train generative models. The attacker aims to reconstruct the data from \textit{both the hacked client and other clients} by shared gradients.}
    \label{fig:overview}
\end{figure}

\textbf{Problem Settings:} In general, we assume that the attacker only has access to the leaked gradients and the model architecture. Our method does not need the parameters of the leaked model or the labels of the training data. As in Figure~\ref{fig:overview}, we adopt a similar premise to DLG~\cite{NEURIPS2019_60a6c400}: the attacker hacks the channel to inject data to one client, which shares the gradient with the server and other clients. The attack aims to reconstruct the data from \textit{both the hacked client and other clients} by shared gradients. When combining with iterative optimization methods, we additionally assume that the attacker can repeatedly query different clients to obtain gradients. The problem settings and comparison with existing literature are summarized in Table~\ref{tab:my_label}.

\textbf{Challenges:} Figure~\ref{fig:overview} provides an overview of the generative approach. Beyond the basic case where the training data for generative model follows the same distribution as the input data for recovery, we may encounter more challenging scenarios: (1) The clients may send defensive inaccurate gradients, like clipped gradients~\cite{NEURIPS2019_60a6c400} and noisy gradients~\cite{abadi2016deep}; (2) When recovering data from other clients based on their shared gradients, challenges arise due to distributional shifts between data on different nodes and slight discrepancies in model parameters across nodes due to lack of synchronization.
Generative methods can adapt to both scenarios, unlike iterative optimization-based methods, which are not applicable for scenario (2) since the attacker has no access to inject data to unhacked clients and is therefore unable to reconstruct training data from them.

Compared with iterative optimization-based methods, our method can train generative models offline and is much more efficient during deployment, making it well-suited for real-time reconstruction tasks with a small tolerated delay.
Compared with latent space–based methods, GIT demonstrates greater robustness to out-of-distribution (OOD) scenarios, as it focuses on inverting the backpropagation process and implicitly recovering the model parameters. 
Compared with other generative model-based methods, GIT is broadly applicable to leaked models with diverse architectures and achieves better performance. 
Moreover, GIT remains effective under challenging reconstruction scenarios.

We summarize our contributions as following points: 
\begin{itemize}

    \item We propose \textbf{\textit{Gradient Inversion Transcript} (GIT)}, which is inspired by back-propagation and constructs a generative model whose architecture is tailored to adapt the leaked model. GIT is shown to effectively reconstruct the input data given its gradient without the knowledge of model parameters and data labels. 
    
    \item GIT can be efficiently deployed after offline training. Compared with existing methods, GIT can achieve the best performance in most cases. In addition, the outputs of GIT can serve as the prior for gradient matching, further improving the performance. 
    
    \item GIT is generally applicable and has robust performance under some challenging performance. It remains effective under discrepancies in model parameters, and achieves best performance under inaccurate gradients and data distributional shift.
\end{itemize}
\section{Related Work}

\begin{table}[H]
    \centering 
    \renewcommand{\arraystretch}{1.3}
    \begin{tabular}{c|ccccccc}
    \Xhline{4\arrayrulewidth}
        \multirow{2}{*}{Method} & Model & Model & Shared & Gradient & Output & Data & Public \\
         & Parameters & Architecture & Gradients & Query & Logit & Label & Data \\
    \hline
         \cite{haim2022reconstructing} & \ding{51} & \ding{51} & \ding{55} & \ding{55} & \ding{55} & \ding{55} & \ding{55} \\
         \cdashline{1-8}
         \citep{phong2017privacy, NEURIPS2019_60a6c400, wei2020framework, geiping2020inverting, wang2020sapag, NEURIPS2021_3b3fff64, chen2024recovering} & \ding{55} & \ding{55} & \ding{51} & \ding{51} & \ding{55} & \ding{51} & \ding{55} \\
         \citep{zhao2020idlg, ma2023instance} & \ding{55} & \ding{55} & \ding{51} & \ding{51} & \ding{55} & dist. & \ding{55} \\
        \citep{dimitrov2024spear} & \ding{51} & \ding{51} & \ding{51} & \ding{55} & \ding{51} & \ding{51} & \ding{55} \\
         \citep{zhu2020r, wang2023reconstructing} & \ding{51} & \ding{51} & \ding{51} & \ding{55} & \ding{55} & \ding{51} & \ding{55} \\
         \cdashline{1-8}
         \citep{yin2021see, wu2025dggi} & \ding{55} & \ding{55} & \ding{51} & \ding{51} & \ding{55} & dist. & \ding{51} \\
          \citep{jeon2021gradient, li2022auditing, fang2023gifd} & \ding{55} & \ding{55} & \ding{51} & \ding{51} & \ding{55} & \ding{51} & \ding{51} \\
         \cdashline{1-8}
         \citep{wu2023learning}, GIT & \ding{55} & \ding{51} & \ding{51} & \ding{51} & \ding{55} & \ding{55} & \ding{51} \\
    \Xhline{4\arrayrulewidth}
    \end{tabular}
    \caption{The comparison in terms of attacker's access for different input data reconstruction methods. \textbf{dist.} means access to distribution of labels. The four categories of methods are separated by dashed lines and, from top to bottom, are: parameter-based methods, iterative optimization-based methods, latent space–based methods, and generative model–based methods.}
    \label{tab:my_label}
\end{table}

To recover the individual training data, gradient-based reconstruction is the most commonly investigated scenario. Nevertheless, it is also worth noting that reconstructing datasets using model parameters only is also viable. 
Such parameter-based methods~\cite{haim2022reconstructing} do not utilize any data-dependent information.
Therefore, the method is unable to recover high-quality data and fails to achieve pixel-wise accuracy.
Consequently, gradient inversion attacks are more widely investigated in the context of the leaked gradients.

\textbf{Optimization-Based Methods.} The feasibility of optimization-based method for data reconstruction from gradient leakage was initially explored by \cite{phong2017privacy}
\cite{NEURIPS2019_60a6c400} demonstrated its practicality by proposing Deep Leakage from Gradients (DLG).
DLG optimizes a randomly generated dummy input to estimate the training data by matching its gradients and the leaked ground truth gradients.
There are several subsequent works improving DLG from optimization perspectives~\cite{wei2020framework, geiping2020inverting, wang2020sapag, zhu2020r, wang2023reconstructing} with settings as shown in Table~\ref{tab:my_label}.
\cite{chen2024recovering} considers a more realistic scenario for data reconstruction, such as multiple local epochs, heterogeneous data and various optimizers.

\textbf{Generative-Based Methods.} Unlike optimization-based methods, generative approaches estimate the distribution of user data using a generative model, which maps the leaked gradient to input data estimation or the initial dummy input for subsequent optimization-based refinement. 
Generative-based methods generally have two major categories which are based on latent space and generative models, respectively.
The first type trains a latent space representation and uses a pre-trained generative model to synthesize estimations of the user data. The second type directly trains a generative model to map leaked gradients to the corresponding user data.

Early attempts of the latent space-based method~\cite{yin2021see} use a pre-trained generative model to produce image priors for reconstruction. Building on this, GIAS~\cite{jeon2021gradient, NEURIPS2021_3b3fff64} employs a generative adversarial network (GAN)~\cite{goodfellow2014generative} as the generative model and alternately searches both the latent space and the parameter space of the generator. 
However, GIAS is computationally prohibitive, as it requires training a new generator for each reconstructed image.
In this context, there are several works~\cite{li2022auditing, fang2023gifd, wu2025dggi} on improving the efficiency and performance of GIAS under different settings as shown in Table~\ref{tab:my_label}.

The generative model-based method was originally proposed by~\cite{wu2023learning} as Learning to Invert (LTI). 
Specifically, they use a three-layer multi-layer perceptron (MLP) with a fixed hidden size as the generative model.
LTI employs a generative model of a fixed architecture regardless of the leaked model, which may not be optimal. In contrast, we introduce gradient inversion transcript (GIT), which is a framework that dynamically selects the architecture of the threat model based on the leaked model to enhance performance.

\section{Input Data Reconstruction by Back-Propagation} \label{sec:analysis}

\subsection{A General Framework} \label{subsec:general_layer}

\begin{wrapfigure}[12]{r}{0.4\textwidth}
\small
\centering
\includegraphics[width = \linewidth]{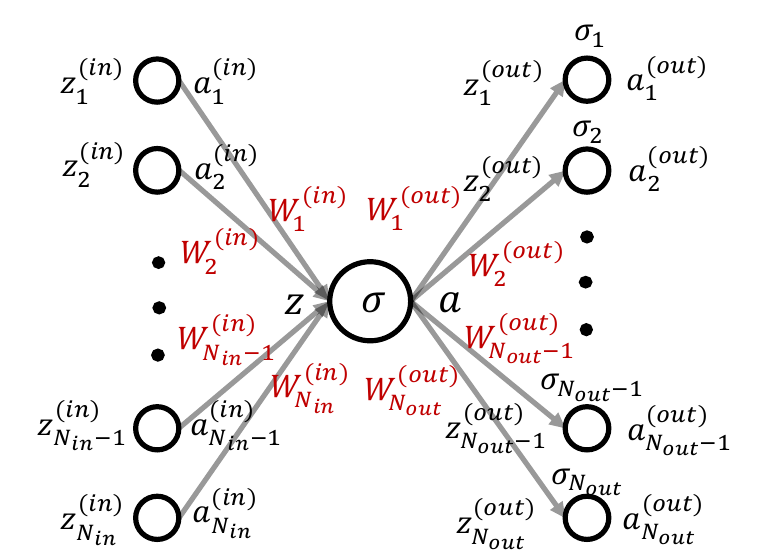}
\caption{An MIMO layer.} \label{fig:mimo}
\end{wrapfigure}
As in Figure~\ref{fig:mimo}, we consider the generic architecture of a multi-input multi-output (MIMO) layer with nonlinear elementwise activation functions as follows:
\begin{equation} \label{eq:nn}
\begin{aligned}
    &\vz = \sum_{i=1}^{N_{in}} \rmW_i^{(in)}\va_i^{(in)},\
    \vz_j^{(out)}=\rmW_j^{(out)} \va, \\ &\va = \sigma(\vz), \ \va_j^{(out)}=\sigma_j(\vz_j^{(out)}),
    \ j=1,...,N_{out}\\
\end{aligned}
\end{equation}

The MIMO layer is connected with $N_{in}$ input layers and $N_{out}$ output layers. $\sigma$ and $\{\sigma_i\}_{i = 1}^{N_{out}}$ are the nonlinear activation functions. We let $B$ be the batch size, $\vz \in \sR^{B \times d}$ and $\va \in \sR^{B \times d}$ represent the pre-activation and post-activation of this MIMO layer, respectively. Similarly, $\left\{\left(\vz_i^{(in)} \in \sR^{B \times d_i^{(in)}}, \va_i^{(in)} \in \sR^{B \times d_i^{(in)}}\right)\right\}_{i = 1}^{N_{in}}$ and $\left\{\left(\vz_j^{(out)} \in \sR^{B \times d_j^{(out)}}, \va_j^{(out)} \in \sR^{B \times d_j^{(out)}}\right)\right\}_{j = 1}^{N_{out}}$ represent the pre-activation and post-activation pairs for the input layers and the output layers, respectively.
In addition, $\left\{\rmW_i^{(in)} \in \sR^{d \times d_i^{(in)}}\right\}_{i = 1}^{N_{in}}$ and $\left\{\rmW_j^{(out)} \in \sR^{d_j^{(out)} \times d}\right\}_{j = 1}^{N_{out}}$ refer to the weights connecting this layer and its adjacent layers. We replace notation $\rmW$ with $\vg$ to represent the gradient of the loss function $\gL$ w.r.t its weights, e.g., $\vg_i^{(in)} = \nabla_{W_i^{(in)}} \gL$, $\vg_j^{(out)} = \nabla_{W_j^{(out)}} \gL$.
We omit the bias term for notation simplicity, since the bias terms can be incorporated as part of the weight matrices.

We have the following equations by back-propagation: 
\begin{equation} \label{eq:grad}
\begin{aligned}
    &\vg_j^{(out)} = \frac{\partial \gL}{\partial \vz_j^{(out)}} \otimes \va^T,\ 
    &\vg_i^{(in)} 
    = \left( \left(\sum_{j=1}^{N_{out}} \rmW_j^{(out)T} \otimes \frac{\partial \gL}{\partial \vz_j^{(out)}} \right) \odot \sigma'(\vz) \right) \otimes \va_i^{(in)T}
\end{aligned}
\end{equation}
Here we define operator $\otimes$ as tensor multiplication and operator $\odot$ as broadcast row-wise product.
In addition, $\vz$, $\va$ are broadcast as a tensor of shape $B \times d \times 1$, similar broadcast mechanisms are applied to $\vz^{(out)}_j$ and $\vz^{(in)}_i$; $\rmW_j^{(out)}$ is broadcast as a tensor of shape $1 \times d^{(out)}_j \times d$, and the transpose operator $(\cdot)^T$ switches the second and the third dimensions of a 3-d tensor.
Based on Equation (\ref{eq:grad}), we cancel out $\partial \gL / \partial \vz_j^{(out)}$ and approximate the input of the layer as follows:
\begin{equation} \label{eq:reconstruct}
\begin{aligned}
    \va_i^{(in)T} \simeq \left( \left( \sum_{j=1}^{N_{out}} \rmW_j^{(out)T} \otimes \vg_j^{(out)} \otimes  (\va^T)^+ \right) \odot \sigma'(\vz) \right)^{+} \otimes \vg_i^{(in)} 
\end{aligned}
\end{equation}
Here we use $(\cdot)^+$ to represent the Moore–Penrose inverse of a matrix.
For a third-order tensor, $(\cdot)^+$ calculate the Moore-Penrose inverse of each of its subspace via the first dimension. Equation~(\ref{eq:reconstruct}) establishes the formulation wherein we leverage the gradients, the parameters and the output activation to estimate the input data of a neuron.
For a neural network of general architecture, we can estimate the input of each layer following back-propagation and ultimately obtain the reconstructed input data.

\textbf{Mini-Batch Training.} In mini-batch training, $\vg_i^{(in)}$ and $\vg^{(out)}_j$ obtained by Equation~(\ref{eq:grad}) contains the gradient information for all data instances in the mini-batch. In practice, the leaked the gradient is their average over the mini-batch, that is $\vg^{(in)}_i \leftarrow \E_b \vg^{(in)}_i[b,:,:]$, $\vg_j^{(out)} \leftarrow \E_b \vg^{(out)}_j[b,:,:]$. Since the leaked gradient is the average over the mini-batch. When reconstructing the input data, we broadcast the leaked gradient in the dimension of batch size in Equation~(\ref{eq:reconstruct}).

\textbf{Generality.} Our analysis is generic and can be applied to general neural network architectures as long as they support back-propagation. For multi-layer perceptrons (MLP) and vanilla convolutional neural networks (CNN) like LeNet, we have $N_{in} = N_{out} = 1$ for all layers; for residual networks (ResNet), we have $N_{out} > 1$ for layers which receive the inputs from both the preceding layer and the shortcut connections. Our framework is also compatible with more complicated architectures like attention mechanism in transformers~\cite{vaswani2017attention}. We defer detailed derivation for these popular architectures in Appendix~\ref{app:architecture}.


\subsection{Modularized Input Data Reconstruction} \label{subsec:module_git}

For models with large amount of parameters, it would be computationally expensive to infer the input data by recursively using Equation~(\ref{eq:reconstruct}). In this context, we formulate the large model as a composition of several modules and apply input data reconstruction on the module level instead of the layer level.
We re-consider the multi-input multi-output (MIMO) layer as in Section~\ref{subsec:general_layer} with input and output connections followed by functions $\left\{f^{(in)}_i\right\}_{i = 1}^{N_{in}}$, $\left\{f^{(out)}_j\right\}_{j = 1}^{N_{out}}$, respectively:
\begin{equation} \label{eq:module}
    \begin{aligned}
        \vz = \sum_{i = 1}^{N_{in}} f_i^{(in)} (\rmW^{(in)}_i \va_i^{(in)}),\ \vz_j^{(out)} = f_j^{(out)}(\rmW^{(out)}_j \va),\ j = 1, ..., N_{out}.
    \end{aligned}
\end{equation}
$\va^{(in)}_i$ and $\va$ are calculated in the same way as in Equation~(\ref{eq:nn}). We follow the derivation as in Section~\ref{subsec:general_layer} and obtain the following equation for modularized input data reconstruction:
\begin{equation} \label{eq:reconstruct_module}
\begin{aligned}
    \va_i^{(in)T} = \left( \left( \sum_{j=1}^{N_{out}} \rmW_j^{(out)T} \otimes \vg_j^{(out)} \otimes  (\va^T)^+ \right) \odot \sigma'(\vz) \otimes f^{'(in)}_i (\rmW^{(in)}_i \va^{(in)}_i) \right)^{+} \otimes \vg_i^{(in)} 
\end{aligned}
\end{equation}

Equation~(\ref{eq:reconstruct_module}) demonstrates modularized input data reconstruction. 
It establishes a high-level formulation to estimate input data for large models.
However, we need the gradient information from the input module $f^{(in)}_i$ to estimate $f^{'(in)}_i (\rmW^{(in)}_i \va^{(in)}_i)$, which will be elaborated in the next section.

\section{GIT: Gradient Inverse Transcript} \label{methodology}

\textbf{Exact-GIT.} Based on Equation~(\ref{eq:reconstruct}) and the analyses in Section~\ref{subsec:general_layer}, the input value $\va_i^{(in)}$ of a general MIMO layer can be estimated from the activation $\va$, the gradient $\vg^{(in)}_i$ of the input weight and output weights $\{\rmW^{(out)}_j\}_{j = 1}^{N_{out}}$. 
Therefore, we can recursively utilize Equation~(\ref{eq:reconstruct}) to reconstruct the input data by a generative model with all unknown variables, such as the weights, as its trainable parameters.
We use the mean square error between the true input data and its estimation as the loss objective function.
Once trained, the generative model can subsequently reconstruct the training data batch using the leaked gradients as input during inference.

The detailed pseudo-code for the training and the inference phase is shown as Algorithm~\ref{alg:general_train}. 
The key innovation of our method is that we adaptively adjust the architectures of the generative models by Equation~(\ref{eq:reconstruct}) based on the leaked model and map the leaked gradients to the estimated input data, so we name it \textit{gradient inverse transcript (GIT)}.
We further name our method \textit{Exact-GIT} when we strictly follow Equation~(\ref{eq:reconstruct}), i.e. using model weights as parameters for the generative models, for all layers to reconstruct the input data.
We present some specific example architectures in Appendix~\ref{app:architecture}. The results of the Exact-GIT implementation are presented in Appendix~\ref{app:exact_git}.
\begin{algorithm}[htbp]
\caption{Training and Inference of GIT}
\label{alg:general_train}
\begin{algorithmic}[1]
\State \underline{\textbf{Training input:}} 
Training set of the generative model, i.e., input-gradient pairs $\gD = \{\left(\vx^{(i)}, \vg^{(i)}\right)\}_{i = 1}^N$. Epoch budget $E$.  Batch size $B$. Learning rate $\eta$.
\State Initialization: Construct the generative model $M$ parameterized by $\Theta$ based on the architecture of the leaked model. Popular architectures are discussed as examples in Appendix~\ref{app:architecture}.
\For{the epoch index from $1$ to $E$}
    \For{the batch index from $1$ to $N / B$}
        \State Sample one mini-batch $\{\left(\vx^{(b_i)}, \vg^{(b_i)}\right)\}_{i = 1}^B$
        \If{use \textit{Exact-GIT}}
            \State Estimate the input $\{\widehat{\vx}^{(b_i)}\}_{i = 1}^B = M(\Theta, \{\vg^{(b_i)}\}_{i = 1}^B)$ by recursively using Equation~(\ref{eq:reconstruct}).
        \Else
            \State Estimate the input $\{\widehat{\vx}^{(b_i)}\}_{i = 1}^B = M(\Theta, \{\vg^{(b_i)}\}_{i = 1}^B)$ by recursively using Equation~(\ref{eq:reconstruct}) or Equation~(\ref{eq:reconstruct_module}) with right hand side replaced by a shallow MLP discussed in~Section~\ref{methodology}.
        \EndIf
        \State Calculate the loss $\gL^{(gen)} = \frac{1}{2B}\sum_{i = 1}^B\|\widehat{\vx}^{b_i} - \vx^{b_i}\|$ and update $\Theta \gets \Theta - \eta \nabla_{\Theta} \gL^{(gen)}$
    \EndFor
\EndFor
\State \underline{\textbf{Training output:}} GIT generator $M$ with learned parameters $\Theta$.
\State 
\State \underline{\textbf{Inference input:}} GIT generator $M$ with parameters $\Theta$. Leaked gradients $\{\vg^{(i)}\}_{i = 1}^{N'}$.
\State \underline{\textbf{Inference output:}} Input data estimation $\{\widehat{\vx}^{(i)}\}_{i = 1}^{N'} = M(\Theta, \{\vg^{(i)}\}_{i = 1}^{N'})$
\end{algorithmic}
\end{algorithm}


\textbf{Coarse-GIT.} Exact-GIT enjoys good interpretability but is computationally expensive for large models.
Moreover, the Moore-Penrose inverse in Equation~(\ref{eq:reconstruct}) would introduces numerical instability issues for large-scale tensors in practice.
To tackle these issues, compared with Equation (\ref{eq:reconstruct}), we can also model such estimation in a more coarse-grained manner and name the corresponding method \textit{Coarse-GIT}.
Specifically, we utilize a shallow multi-layer perceptron (MLP) $m_\theta$, parameterized by $\theta$, to approximate the right-hand-side of Equation~(\ref{eq:reconstruct}).
The inputs of this shallow MLP are all the known variables on the right-hand-side of Equation~(\ref{eq:reconstruct}), including the leaked gradients and the output activation.
Therefore, like Equation~(\ref{eq:reconstruct}), Coarse-GIT recursively estimates each layer's input by $\va^{(in)}_i = m_\theta \left(\{\vg_j^{(out)}\}_{j = 1}^{N_{out}}, \vg_i^{(in)}, \va \right)$. 
The generative model comprises multiple shallow MLPs, with orders based on back-propagation and collectively trained to minimize the difference between the estimated input and the corresponding ground truth.

Coarse-GIT also supports modularized input data reconstruction as discussed in Section~\ref{subsec:module_git}, which is more computationally affordable. 
It employs a shallow MLP $m_\theta$ to estimate the right-hand-side of Equation~(\ref{eq:reconstruct_module}): $\va^{(in)}_i = m_\theta\left(\{\vg_j^{(out)}\}_{j = 1}^{N_{out}}, \vg^{f{'(in)}}_i, \vg_i^{(in)}, \va \right)$ where $\vg^{f' (in)}_i$ represents the leaked gradients for the parameters in the input module $f^{(in)}_i$.
Compared with layerwise reconstruction, modularized reconstruction only considers the high-level topologies of the leaked model, making it suitable for large models.

\textbf{Bootstrap.} For both Exact-GIT and Coarse-GIT, we need to estimate the output logits, i.e., last layer's output, to start the recursive estimation. The average output logits over the mini-batch can be analytically estimated if the last layer has a bias term~\cite{zhu2020r}. Otherwise, we use the leaked gradients for the weight of the last layers to estimate the output logits by a shallow MLP. The ablation studies are presented in Appendix~\ref{app:anlysis}.


\section{Experiments} \label{sec:exp}

We comprehensively assess our methods on various datasets, including CIFAR-10~\citep{krizhevsky2009learning}, ImageNet~\citep{deng2009imagenet} and facial datasets (Facial Expression Recognition (FER) from kaggle, Japanese Female Facial Expression (Jaffe)~\citep{lyons1998japanese}).
Correspondingly, we employ various model architectures, including LeNet~\citep{lecun1998gradient}, ResNet~\citep{he2016deep} and ViT~\citep{dosovitskiy2020image} to comprehensively demonstrate the effectiveness of our methods.
Since the generative models are trained by minimizing the mean square error (MSE) between the ground-truth and the estimated inputs, in addition to MSE, we also use peak signal-to-noise ratio (PSNR), structural similarity index (SSIM), learned perceptual image patch similarity (LPIPS) as metrics to quantitatively and comprehensively evaluate the performance of training data reconstruction. PSNR, SSIM and LPIPS reflect more perceptual and structural differences than MSE.

\subsection{Comparison with Baselines}
\label{sec:baselines}

We compare our method with baselines under two different settings: (1) we directly employ generative models to map the leaked gradient to the reconstructed input data; (2) we first employ generative models to obtain the input data estimation as priors and then refine the estimation by optimization-based methods.
Unless specified, we use $10000$ random samples and their gradients to train the generative models.
More implementation details are deferred to Appendix~\ref{app:experiment_setting}.

\subsubsection{Direct Inference by Generative Models} \label{subsec:direct}

As shown in Table~\ref{table:direct}, we first compare GIT with other generative models in direct inference. Specifically, we compare GIT with Learning to Invert (LTI)~\cite{wu2023learning}, which employs an MLP with approximately the same number of parameters as the generators.
In addition, we include the performance of optimization-based methods for reference, such as Deep Leakage from Gradients (DLG)~\cite{NEURIPS2019_60a6c400} and Inverting Gradients (IG)~\cite{geiping2020inverting}.
The computational overhead for optimization-based methods and generative methods are fundamentally different.
Optimization-based methods necessitate a complete optimization process for each batch data recovery, whereas generative methods need to train a generative model capable of retrieving data from the corresponding leaked gradients. We run both types of methods until their convergence and report the training and inference time for comparison.

The results in Table~\ref{table:direct} include different tasks and network architectures.
To save memory consumption and guarantee numerical stability, we adopt Coarse-GIT for all architectures and specifically modularized reconstruction for ViT. The GIT implementation details for specific architectures are deferred to Appendix~\ref{app:experiment_setting}.
The results indicate that GIT outperforms in most cases and metrics than baselines, including both optimization-based methods and generative methods.

Visual inspection of reconstructed ImageNet samples by GIT as shown in Appendix~\ref{app:visual} reveals that images with large uniform color regions tend to be recovered more accurately, while those containing complex structures or multiple objects exhibit inferior reconstruction quality. This is consistent with the observations in Table~\ref{table:direct} that GIT always performs the best in term of MSE but may underperform in term of LPIPS which focuses more on the image structure. Therefore, instead of directly employing GIT for inference, we further utilize it as an image prior to guide optimization-based methods toward more perceptually accurate results.
\begin{table}[!ht]
    \centering
    \caption{Quantitative comparison for different datasets and models in terms of different metrics. Dashed lines separate generative-based methods with optimization-based ones. The training time represents the time cost for training the generative model. The inference time represents the average time to reconstruct one input data instances from the leakage gradients during inference.}
    \renewcommand{\arraystretch}{1.2}
    \begin{tabular}{l!{\vrule width 1.1pt}c!{\vrule width 1.1pt}c!{\vrule width 1.1pt}cccc!{\vrule width 1.1pt}cc}
        \Xhline{4\arrayrulewidth}
        Dataset & \begin{tabular}{l} Leaked \\ Model \end{tabular} & Method & MSE$\downarrow$ & PSNR$\uparrow$ & LPIPS$\downarrow$ & SSIM$\uparrow$ & \begin{tabular}{c} Training \\ Time (s) \end{tabular} & \begin{tabular}{c} Inference \\ Time (s) \end{tabular} \\ \Xhline{4\arrayrulewidth}
        \multirow{8}{*}{CIFAR10} 
        & \multirow{4}{*}{LeNet}  & DLG   & 0.073 & 11.32 & 0.2380 & 0.0847 & / & 1660 \\
        &                         & IG    & 0.082 & 11.27 & 0.3916 & 0.1193 & / & 1899 \\
        \cdashline{3-9}
        &                         & LTI   & 0.015 & 19.17 & \textbf{0.2202} & 0.5304 & 8549 & 0.0030 \\
        &                         & GIT   & \textbf{0.010} & \textbf{20.38} & 0.2663 & \textbf{0.5533} & 8071 & 0.0025 \\
        \Xcline{2-9}{0.9pt} 
        & \multirow{4}{*}{ResNet} & DLG   & 0.084 & 10.93 & 0.3813 & 0.0667 & / & 7474 \\
        &                         & IG    & 0.080 & 10.75 & \textbf{0.2489} & 0.0739 & / & 6875 \\
        \cdashline{3-9}
        &                         & LTI   & 0.035 & 15.32 & 0.4400 & 0.2888 & 5212 & 0.0020 \\
        &                         & GIT   & \textbf{0.032} & \textbf{15.53} & 0.3957 & \textbf{0.3188} & 4019 & 0.0013 \\ 
        \Xhline{3\arrayrulewidth}
        \multirow{8}{*}{ImageNet} 
        & \multirow{4}{*}{ResNet} & DLG   & 0.147 & 9.25 & 0.8754 & 0.1324 & / & 3974 \\
        &                         & IG    & 0.161 & 9.17 & 0.8802 & 0.1283 & / & 4103 \\
        \cdashline{3-9}
        &                         & LTI   & 0.043 & 14.25 & 0.9017 & 0.3418 & 10200 & 0.0007 \\
        &                         & GIT   & \textbf{0.039} & \textbf{14.42} & \textbf{0.8513} & \textbf{0.3507} & 13011 & 0.0008 \\
        \Xcline{2-9}{0.9pt} 
        & \multirow{4}{*}{ViT}    & DLG   & 0.172 & 7.57 & 0.9513 & 0.1217 & / & 3734 \\
        &                         & IG    & 0.175 & 7.64 & 0.9427 & 0.1210 & / & 3025 \\
        \cdashline{3-9}
        &                         & LTI   & 0.046 & 13.37 & 0.9223 & 0.2117 & 9738 & 0.0029 \\
        &                         & GIT   & \textbf{0.034} & \textbf{15.25} & \textbf{0.8365} & \textbf{0.3774} & 6717 & 0.0025 \\
        \Xhline{3\arrayrulewidth}
    \end{tabular}
    \label{table:direct}
\end{table}

\subsubsection{Optimization-Based Data Reconstruction Using Generative Models as Priors}
\label{subsec:hybrid}

Optimization-based methods are shown highly sensitive to the initialization of dummy inputs~\cite{wei2020framework}.
Therefore, recent methods, such as Gradient Inversion with Generative Image Prior (GIAS)~\cite{jeon2021gradient}, propose to utilize generative models to generate image priors as initialization of optimization-based methods. 
In this context, we can employ GIT to generate informative priors, which are subsequently refined through iterative optimization-based methods like IG. 
For generative methods~\citep{yin2021see, jeon2021gradient, li2022auditing, fang2023gifd, wu2025dggi, wu2023learning}, we select GIAS~\cite{jeon2021gradient} and LTI~\cite{wu2023learning} as a representative baseline for comparison. To ensure fairness, all generative models are trained from scratch without relying on any pretraining, and are followed by the same optimization-based method.


As shown in Table~\ref{table:prior}, using GIT to generate priors and refine the reconstructed image by IG (GIT+IG) have the best performance in almost all cases and all metrics.
In addition, GIT+IG always has better performance than LTI+IG, indicating GIT based on adaptive architectures can provide better priors than LTI based on fixed architectures.
Furthermore, GIT+IG, as a hybrid approach, not only converges faster but also outperforms both GIT and IG when used individually, demonstrating its superior effectiveness.
We visualize the convergence curves of IG with and without the image prior in Appendix~\ref{app:curve}, highlighting the benefits of incorporating generative priors into the optimization process.

\begin{table}[!ht]
    \centering
    \caption{Quantitative comparison for different datasets and models in terms of different metrics. The performance of IG is used as references. The training time represents the time cost for training the generative model. The inference time represents the average time to reconstruct one input data instances from the leakage gradients during inference.}
    \renewcommand{\arraystretch}{1.2}
    \begin{tabular}{l!{\vrule width 1.1pt}c!{\vrule width 1.1pt}c!{\vrule width 1.1pt}cccc!{\vrule width 1.1pt}cc}
        \Xhline{4\arrayrulewidth}
        Dataset & \begin{tabular}{l} Leaked \\ Model \end{tabular} & Method & MSE$\downarrow$ & PSNR$\uparrow$ & LPIPS$\downarrow$ & SSIM$\uparrow$ & \begin{tabular}{c} Training \\ Time (s) \end{tabular} & \begin{tabular}{c} Inference \\ Time (s) \end{tabular} \\ \Xhline{4\arrayrulewidth}
        \multirow{8}{*}{CIFAR10} 
        & \multirow{4}{*}{LeNet}  & IG  & 0.082 & 11.27 & 0.3916 & 0.1193 & / & 1899 \\
        \cdashline{3-9}
        &                         & GIAS+IG  & 0.009 & 21.45 & 0.0328 & 0.8925 & 10025 & 242 \\
        &                         & LTI+IG  & 0.002 & 30.86 & 0.0025 & 09356 & 8549 & 158 \\
        &                         & GIT+IG  & \textbf{0.001} & \textbf{31.25} & \textbf{0.0009} & \textbf{0.9551} & 8071 & 161 \\
        \Xcline{2-9}{0.9pt} 
        & \multirow{4}{*}{ResNet} & IG  & 0.080 & 10.75 & 0.2489 & 0.0739 & / & 6875 \\
        \cdashline{3-9}
        &                         & GIAS+IG    & 0.019 & 18.96 & 0.2437 & 0.6125 & 10892 & 187 \\ 
        &                         & LTI+IG  & 0.009 & 20.48 & 0.0092 & 0.8266 & 5212 & 1651 \\
        &                         & GIT+IG  & \textbf{0.002} & \textbf{31.34} & \textbf{0.0041} & \textbf{0.9218} & 4019 & 1655 \\
        \Xhline{3\arrayrulewidth}
        \multirow{8}{*}{ImageNet} 
        & \multirow{4}{*}{ResNet} & IG  & 0.161 & 9.17 & 0.8802 & 0.1283 & / & 4103 \\
        \cdashline{3-9}
        &                         & GIAS+IG & 0.037 & 14.32 & 0.8218 & 0.3765 & 27453 & 3209 \\
        &                         & LTI+IG  & 0.029 & 15.38 & 0.7434 & 0.4129 & 10200 & 2065 \\
        &                         & GIT+IG  & \textbf{0.021} & \textbf{16.78} & \textbf{0.6995} & \textbf{0.4758} & 13011 & 1998 \\
        \Xcline{2-9}{0.9pt} 
        & \multirow{4}{*}{ViT}    & IG  & 0.175 & 7.64 & 0.9427 & 0.1210 & / & 3025  \\
        \cdashline{3-9}
        &                         & GIAS+IG & 0.039 & 14.09 & 0.7572 & \textbf{0.5239} & 36950 & 3997 \\
        &                         & LTI+IG  & 0.029 & 15.96 & 0.7250 & 0.4231 & 6138 & 2950 \\
        &                         & GIT+IG  & \textbf{0.019} & \textbf{17.21} & \textbf{0.6730} & 0.5025 & 6717 & 2987 \\
        \Xhline{3\arrayrulewidth}
    \end{tabular}
    \label{table:prior}
\end{table}

\subsection{Reconstruction Under Challenging Situations}

In this section, we investigate the robustness of reconstruction methods under different challenging situations, including inaccurate leaked gradients and the substantial distributional shift between the public data and the training data.
In such situations, optimization-based methods are not applicable or do not have competitive performance.
Therefore, we mainly compare the results from the direct inference by generative models. More details are deferred to Appendix~\ref{app:experiment_setting}.

\subsubsection{Inaccurate Gradients}

Prior works~\cite{wu2023learning} have shown degraded performance of optimization-based methods when the leaked gradients are inaccurate. Unlike optimization-based methods which observe significant performance degradation in the case of inaccurate gradients, Appendix~\ref{app:prune} shows that generative methods primarily utilize the gradient elements with large absolute values to generate outputs, indicating robustness in such challenging cases.

In Table~\ref{table:noise}, we consider the leaked gradients perturbed by isotropic Gaussian noise with different standard deviation (std). We compare different generative methods and include optimization-based methods like IG for reference.
The results confirm the vulnerability of optimization-based methods against gradient perturbations.
Among generative methods, GIT performs the best in all cases and all metrics, showing minimal susceptibility to inaccurate gradients.
\begin{table}[h]
\centering
\caption{Comparison of metrics under gradient perturbation with varying noise variance. The batch size is fixed at $1$, and the leaked model is LeNet with 5 layers.} 
\renewcommand{\arraystretch}{1.2}
\begin{tabular}{c!{\vrule width 1.1pt}c!{\vrule width 1.1pt}cccc}
    \Xhline{4\arrayrulewidth}
    Method & std of noise & MSE$\downarrow$ & PSNR$\uparrow$ & LPIPS$\downarrow$ & SSIM$\uparrow$ \\ 
    \Xcline{1-6}{0.9pt}
    \multirow{3}{*}{IG}  & None   & 0.082 & 11.27 & 0.3916 & 0.1193 \\
                          & 0.01  & 0.105 & 9.79 & 0.4098 & 0.1172 \\
                          & 0.1   & 0.162 & 9.18 & 0.4320 & 0.1126 \\
    \Xcline{1-6}{0.9pt}
    \multirow{3}{*}{LTI}  & None  & 0.015 & 19.17 & 0.2202 & 0.5304 \\
                          & 0.01  & 0.015 & 19.16 & 0.2205 & 0.5300 \\
                          & 0.1   & 0.015 & 19.16 & 0.2199 & 0.5287 \\
    \Xcline{1-6}{0.9pt}
    \multirow{3}{*}{GIAS}  & None  & 0.012 & 19.21 & 0.2350 & 0.5398 \\
                          & 0.01  & 0.012 & 19.18 & 0.3010 & 0.5219 \\
                          & 0.1   & 0.013 & 18.87 & 0.3113 & 0.5189 \\
    \Xcline{1-6}{0.9pt}
    \multirow{3}{*}{GIT}  & None  & 0.010 & 20.38 & 0.2663 & 0.5533 \\
                          & 0.01  & 0.010 & 20.36 & 0.2675 & 0.5520 \\
                          & 0.1   & 0.010 & 20.37 & 0.2669 & 0.5522 \\
    \Xhline{3\arrayrulewidth}
\end{tabular}
\label{table:noise}
\end{table}

\subsubsection{Distribution Shift}

As shown in Figure \ref{fig:overview}, GIT is trained on the public dataset injected by the attacker, and aims to reconstruct the local dataset. There may be a distributional shift between these two datasets, which could influence the effectiveness of the generative attack model.

We consider two possible scenarios of distribution shifts: (1) the public and local datasets come from different subsets of the same dataset, with distribution differences arising from overlapping but distinct classes; 
(2) the public datasets are subsets of huge but more general datasets, such as FER, while the local datasets held by individual clients are more specific ones, such as Jaffe.

Our experiment in Table~\ref{table:shift} investigate both scenarios above. For CIFAR-10 and ImageNet, the public data and the local data share $6$ classes and the rest classes are distinct. For facial dataset, the public data and the local data have different resolutions and significant distributional shifts.
The results in Table~\ref{table:shift} indicate that GIT demonstrates the strongest generalization ability across distribution shifts and achieves the best performance on the local dataset. 
GIT learns an implicit representation of the leaked model's parameters by its adaptive architecture, which is more agnostic to the data distribution.
By contrast, GIAS learns a latent space that captures the distribution characteristics of the public dataset, which requires the public and local datasets to share highly similar features to perform well.
\begin{table}[H]
    \centering
    \caption{Comparison of the metrics under distributional shift. For each dataset, we select a federated learning model architecture that is well-suited for its classification task: LeNet is used for CIFAR-10, ResNet for ImageNet, and Vision Transformer (ViT) for facial datasets. The "classes" in the public data and local data represent categories sampled to form datasets. In the case of facial data, we conduct experiment where both the public data and local data come from FER, which serves as a comparison.}
    \renewcommand{\arraystretch}{1.2}
        \begin{tabular}{l!{\vrule width 1.1pt}c!{\vrule width 1.1pt}c!{\vrule width 1.1pt}c!{\vrule width 1.1pt}cccc}
        \Xhline{4\arrayrulewidth}
        Dataset & \begin{tabular}{l} Public \\ Data \end{tabular} & \begin{tabular}{l} Local \\ Data \end{tabular} & Method & MSE$\downarrow$ & PSNR$\uparrow$ & LPIPS$\downarrow$ & SSIM$\uparrow$
        \\ \Xhline{4\arrayrulewidth}
        \multirow{3}{*}{CIFAR10} 
        & \multirow{3}{*}{classes 1-8}    & \multirow{3}{*}{classes 3-10}  
                                  & GIAS    & 0.065 & 11.87 & 0.3670 & 0.3092 \\
        & &                       & LTI     & 0.029 & 15.38 & 0.3028 & 0.3790 \\
        & &                       & GIT     & \textbf{0.020} & \textbf{17.44} & \textbf{0.2155} & \textbf{0.4150} \\
        \Xcline{1-8}{0.9pt} 
        \multirow{3}{*}{ImageNet100} 
        & \multirow{3}{*}{classes 1-53}    & \multirow{3}{*}{classes 48-100}  
                                  & GIAS    & 0.061 & 12.15 & 0.9518 & 0.3126 \\
        & &                       & LTI     & 0.049 & 13.10 & 0.9274 & 0.3150 \\
        & &                       & GIT     & \textbf{0.043} & \textbf{13.67} & \textbf{0.9044} & \textbf{0.3224} \\
        \Xcline{1-8}{0.9pt} 
        \multirow{6}{*}{Facial Data} 
        & \multirow{6}{*}{FER}    & \multirow{3}{*}{FER}  
                                  & GIAS    & 0.020 & 17.73 & 0.4174 & 0.4051 \\
        & &                       & LTI     & 0.020 & 17.60 & 0.4420 & 0.3949 \\
        & &                       & GIT     & \textbf{0.018} & \textbf{17.93} & \textbf{0.3405} & \textbf{0.4228} \\
        \cdashline{3-8}
        &                         & \multirow{3}{*}{Jaffe}  
                                  & GIAS    & 0.042 & 13.77 & 0.5128 & 0.2826 \\
        & &                       & LTI     & 0.033 & 15.21 & 0.4625 & 0.3187 \\
        & &                       & GIT     & \textbf{0.030} & \textbf{15.54} & \textbf{0.3461} & \textbf{0.3244} \\
        \Xcline{2-8}{0.9pt} 

        \Xhline{3\arrayrulewidth}
    \end{tabular}
    \label{table:shift}
\end{table}
\vspace{-0.3cm}

\subsubsection{Discrepancies in Model Parameters}

In federated learning, gradient sharing may be asynchronous~\cite{geiping2020inverting}, leading to slight discrepancies in model parameters across different nodes. Such inconsistencies can affect the performance of both optimization-based and generative reconstruction methods.

To create discrepancies in model parameters, we train each node with different volume of local dataset for several epochs, then we use generative models trained on input-gradient pairs from one node, i.e., the node under attack, to reconstruct the input data from another node, which is not necessarily under attack and has parameter discrepancies.
In this context, larger local data and more training epochs lead to larger parameter discrepancies, making the input data reconstruction task more challenging.

Our experimental results under different settings are summarized in Table~\ref{table:parameter_robust}. We can clearly observe that GIT achieves the best performance under significant parameter discrepancies and is still capable of reconstructing high-quality training data, demonstrating its robustness to variations in model parameters across nodes. 
We also find that increasing the number of local datasets sometimes leads to reduced performance degradation. This may be because a larger number of local datasets increases the likelihood of including samples from classes that are easier to reconstruct, thus introducing a degree of randomness that favors recovery.

\begin{table}[H]
\centering
\caption{Comparison of the MSE for GIT with varying parameter discrepancies. The parameter discrepancies is quantified by volume of local dataset \& number of locally trained epochs. The leaked model for CIFAR10 is LeNet, and for ImageNet is ResNet.} 
\renewcommand{\arraystretch}{1.2}
\begin{tabular}{c!{\vrule width 1.1pt}c!{\vrule width 1.1pt}c!{\vrule width 1.1pt}p{1.5cm}<{\centering}p{1.5cm}<{\centering}p{1.5cm}<{\centering}}
\Xhline{4\arrayrulewidth}
\multirow{2}{*}{Dataset} & Volume of & \multirow{2}{*}{Method} & \multicolumn{3}{c}{Number of Locally Trained Epochs} \\
& Local Dataset & & 0 & 10 & 20 \\
\Xhline{4\arrayrulewidth}
\multirow{9}{*}{CIFAR10} & \multirow{3}{*}{500} & IG & 0.082 & 0.089 & 0.096 \\
& & LTI & 0.015 & 0.019 & 0.023 \\
& & GIT & \textbf{0.010} & \textbf{0.013} & \textbf{0.017} \\
\Xcline{2-6}{0.9pt}
& \multirow{3}{*}{1000} & IG & 0.082 & 0.097 & 0.102 \\
& & LTI & 0.015 & 0.026 & 0.030 \\
& & GIT & \textbf{0.010} & \textbf{0.017} & \textbf{0.020} \\
\Xcline{2-6}{0.9pt}
& \multirow{3}{*}{10000} & IG & 0.082 & 0.096 & 0.107 \\
& & LTI & 0.015 & 0.032 & 0.036 \\
& & GIT & \textbf{0.010} & \textbf{0.029} & \textbf{0.034} \\
\Xhline{2\arrayrulewidth}
\multirow{9}{*}{ImageNet} & \multirow{3}{*}{500} & IG & 0.161 & 0.162 & 0.162 \\
& & LTI & 0.043 & 0.049 & 0.053 \\
& & GIT & \textbf{0.039} & \textbf{0.043} & \textbf{0.044} \\
\Xcline{2-6}{0.9pt}
& \multirow{3}{*}{1000} & IG & 0.161 & 0.162 & 0.163 \\
& & LTI & 0.043 & 0.049 & 0.053 \\
& & GIT & \textbf{0.039} & \textbf{0.043} & \textbf{0.043} \\
\Xcline{2-6}{0.9pt}
& \multirow{3}{*}{10000} & IG & 0.161 & 0.164 & 0.164 \\
& & LTI & 0.043 & 0.048 & 0.051 \\
& & GIT & \textbf{0.039} & \textbf{0.040} & \textbf{0.040} \\
\Xhline{4\arrayrulewidth}
\end{tabular}
\label{table:parameter_robust}
\end{table}

\subsection{Ablation Studies}

\begin{wraptable}{r}{0.45\textwidth}
\centering
\caption{MSE comparison with varying volumes of training data. The dataset is CIFAR10 and the leaked model is LeNet.}
\renewcommand{\arraystretch}{1.2}
\begin{tabular}{c|ccc}
    \Xhline{4\arrayrulewidth}
    \textbf{Data Volume}  & \textbf{LTI} & \textbf{GIAS} & \textbf{GIT}\\
    \hline
    1000  & 0.039 & 0.049 & \textbf{0.027} \\
    2000  & 0.033 & 0.045 & \textbf{0.021} \\
    5000  & 0.027 & 0.036 & \textbf{0.015} \\
    10000  & 0.015 & 0.027 & \textbf{0.010} \\
    \Xhline{4\arrayrulewidth}
\end{tabular}
\label{table:volume}
\end{wraptable}
Training data volume refers to the size of the auxiliary dataset sampled by the attacker from public data. A larger training dataset, akin to performing data augmentation, can enhance the generalization ability of the generative model. 
However, it also incurs higher computational costs, requiring more time and resources to train the model. Moreover, in practice, acquiring a large volume of data with a distribution similar to the local dataset can be challenging.
Given this tradeoff, it is essential to evaluate the performance of the generative approach under different training data volumes.
In this section, we evaluate the performance of GIT using varying amounts of training data: $1,000$, $2,000$, $5,000$ and $10,000$ samples. Table~\ref{table:volume} presents the impact of training data volume on the performance of the generative approach. It demonstrates that even with only $1,000$ input-gradient pairs, GIT is capable of reconstructing reasonable images, indicating that effective recovery is achievable with a limited amount of training data. As the generative model achieves near-perfect performance with larger training sets, we can conclude that increased data volume helps mitigate overfitting and consequently improves model performance. Moreover, GIT consistently outperforms both LTI and GIAS across all settings.

More analyses and ablation studies are deferred to Appendix~\ref{app:anlysis}.

\section{Conclusions}

This work introduces \textit{Generative Gradient Inversion Transcript (GIT)}, a novel method for reconstructing training data in federated learning by exploiting gradient leakage. We propose a reconstruction framework based on generative models, leveraging an adaptive structure inspired by the inverse of backpropagation. Our attack model realistically assumes that an attacker can inject data into a compromised client, enabling the reconstruction of local datasets from both compromised and non-compromised clients. This setup aligns with practical adversarial scenarios in federated learning.
GIT offers a significant efficiency advantage, being more cost-effective in inference than optimization-based methods, and can be seamlessly deployed after offline training. Compared to existing methods, GIT achieves superior performance in most cases. Furthermore, GIT-generated outputs can serve as priors for optimization-based gradient matching approaches, further enhancing attack effectiveness.
GIT demonstrates strong generalization and robustness under challenging conditions, including inaccurate gradients, distributional shifts, and discrepancies in model parameters across clients.

\bibliography{reference}

\begin{thebibliography}{10}

\bibitem{jochems2016distributed}
Arthur Jochems, Timo~M Deist, Johan Van~Soest, Michael Eble, Paul Bulens, Philippe Coucke, Wim Dries, Philippe Lambin, and Andre Dekker.
\newblock Distributed learning: developing a predictive model based on data from multiple hospitals without data leaving the hospital--a real life proof of concept.
\newblock {\em Radiotherapy and Oncology}, 121(3):459--467, 2016.

\bibitem{mcmahan2017communication}
Brendan McMahan, Eider Moore, Daniel Ramage, Seth Hampson, and Blaise~Aguera y~Arcas.
\newblock Communication-efficient learning of deep networks from decentralized data.
\newblock In {\em Artificial intelligence and statistics}, pages 1273--1282. PMLR, 2017.

\bibitem{yang2019ffd}
Wensi Yang, Yuhang Zhang, Kejiang Ye, Li~Li, and Cheng-Zhong Xu.
\newblock Ffd: A federated learning based method for credit card fraud detection.
\newblock In {\em Big Data--BigData 2019: 8th International Congress, Held as Part of the Services Conference Federation, SCF 2019, San Diego, CA, USA, June 25--30, 2019, Proceedings 8}, pages 18--32. Springer, 2019.

\bibitem{NEURIPS2021_3b3fff64}
Yangsibo Huang, Samyak Gupta, Zhao Song, Kai Li, and Sanjeev Arora.
\newblock Evaluating gradient inversion attacks and defenses in federated learning.
\newblock In M.~Ranzato, A.~Beygelzimer, Y.~Dauphin, P.S. Liang, and J.~Wortman Vaughan, editors, {\em Advances in Neural Information Processing Systems}, volume~34, pages 7232--7241. Curran Associates, Inc., 2021.

\bibitem{phong2017privacy}
Le~Trieu Phong, Yoshinori Aono, Takuya Hayashi, Lihua Wang, and Shiho Moriai.
\newblock Privacy-preserving deep learning: Revisited and enhanced.
\newblock In {\em Applications and Techniques in Information Security: 8th International Conference, ATIS 2017, Auckland, New Zealand, July 6--7, 2017, Proceedings}, pages 100--110. Springer, 2017.

\bibitem{NEURIPS2019_60a6c400}
Ligeng Zhu, Zhijian Liu, and Song Han.
\newblock Deep leakage from gradients.
\newblock In H.~Wallach, H.~Larochelle, A.~Beygelzimer, F.~d\textquotesingle Alch\'{e}-Buc, E.~Fox, and R.~Garnett, editors, {\em Advances in Neural Information Processing Systems}, volume~32. Curran Associates, Inc., 2019.

\bibitem{zhao2020idlg}
Bo~Zhao, Konda~Reddy Mopuri, and Hakan Bilen.
\newblock idlg: Improved deep leakage from gradients.
\newblock {\em arXiv preprint arXiv:2001.02610}, 2020.

\bibitem{wei2020framework}
Wenqi Wei, Ling Liu, Margaret Loper, Ka-Ho Chow, Mehmet~Emre Gursoy, Stacey Truex, and Yanzhao Wu.
\newblock A framework for evaluating gradient leakage attacks in federated learning.
\newblock {\em arXiv preprint arXiv:2004.10397}, 2020.

\bibitem{geiping2020inverting}
Jonas Geiping, Hartmut Bauermeister, Hannah Dr{\"o}ge, and Michael Moeller.
\newblock Inverting gradients-how easy is it to break privacy in federated learning?
\newblock {\em Advances in neural information processing systems}, 33:16937--16947, 2020.

\bibitem{zhu2020r}
Junyi Zhu and Matthew Blaschko.
\newblock R-gap: Recursive gradient attack on privacy.
\newblock {\em arXiv preprint arXiv:2010.07733}, 2020.

\bibitem{wang2020sapag}
Yijue Wang, Jieren Deng, Dan Guo, Chenghong Wang, Xianrui Meng, Hang Liu, Caiwen Ding, and Sanguthevar Rajasekaran.
\newblock Sapag: A self-adaptive privacy attack from gradients.
\newblock {\em arXiv preprint arXiv:2009.06228}, 2020.

\bibitem{yin2021see}
Hongxu Yin, Arun Mallya, Arash Vahdat, Jose~M Alvarez, Jan Kautz, and Pavlo Molchanov.
\newblock See through gradients: Image batch recovery via gradinversion.
\newblock In {\em Proceedings of the IEEE/CVF conference on computer vision and pattern recognition}, pages 16337--16346, 2021.

\bibitem{wang2023reconstructing}
Zihan Wang, Jason Lee, and Qi~Lei.
\newblock Reconstructing training data from model gradient, provably.
\newblock In {\em International Conference on Artificial Intelligence and Statistics}, pages 6595--6612. PMLR, 2023.

\bibitem{jeon2021gradient}
Jinwoo Jeon, Kangwook Lee, Sewoong Oh, Jungseul Ok, et~al.
\newblock Gradient inversion with generative image prior.
\newblock {\em Advances in neural information processing systems}, 34:29898--29908, 2021.

\bibitem{li2022auditing}
Zhuohang Li, Jiaxin Zhang, Luyang Liu, and Jian Liu.
\newblock Auditing privacy defenses in federated learning via generative gradient leakage.
\newblock In {\em Proceedings of the IEEE/CVF Conference on Computer Vision and Pattern Recognition}, pages 10132--10142, 2022.

\bibitem{fang2023gifd}
Hao Fang, Bin Chen, Xuan Wang, Zhi Wang, and Shu-Tao Xia.
\newblock Gifd: A generative gradient inversion method with feature domain optimization.
\newblock In {\em Proceedings of the IEEE/CVF International Conference on Computer Vision}, pages 4967--4976, 2023.

\bibitem{wu2023learning}
Ruihan Wu, Xiangyu Chen, Chuan Guo, and Kilian~Q Weinberger.
\newblock Learning to invert: Simple adaptive attacks for gradient inversion in federated learning.
\newblock In {\em Uncertainty in Artificial Intelligence}, pages 2293--2303. PMLR, 2023.

\bibitem{chen2024recovering}
Huancheng Chen and Haris Vikalo.
\newblock Recovering labels from local updates in federated learning.
\newblock {\em arXiv preprint arXiv:2405.00955}, 2024.

\bibitem{wu2025dggi}
Liwen Wu, Zhizhi Liu, Bin Pu, Kang Wei, Hangcheng Cao, and Shaowen Yao.
\newblock Dggi: Deep generative gradient inversion with diffusion model.
\newblock {\em Information Fusion}, 113:102620, 2025.

\bibitem{rosenblatt1958perceptron}
Frank Rosenblatt.
\newblock The perceptron: a probabilistic model for information storage and organization in the brain.
\newblock {\em Psychological review}, 65(6):386, 1958.

\bibitem{abadi2016deep}
Martin Abadi, Andy Chu, Ian Goodfellow, H~Brendan McMahan, Ilya Mironov, Kunal Talwar, and Li~Zhang.
\newblock Deep learning with differential privacy.
\newblock In {\em Proceedings of the 2016 ACM SIGSAC conference on computer and communications security}, pages 308--318, 2016.

\bibitem{haim2022reconstructing}
Niv Haim, Gal Vardi, Gilad Yehudai, Ohad Shamir, and Michal Irani.
\newblock Reconstructing training data from trained neural networks.
\newblock {\em Advances in Neural Information Processing Systems}, 35:22911--22924, 2022.

\bibitem{ma2023instance}
Kailang Ma, Yu~Sun, Jian Cui, Dawei Li, Zhenyu Guan, and Jianwei Liu.
\newblock Instance-wise batch label restoration via gradients in federated learning.
\newblock In {\em The Eleventh International Conference on Learning Representations}, 2023.

\bibitem{dimitrov2024spear}
Dimitar~I Dimitrov, Maximilian Baader, Mark M{\"u}ller, and Martin Vechev.
\newblock Spear: Exact gradient inversion of batches in federated learning.
\newblock {\em Advances in Neural Information Processing Systems}, 37:106768--106799, 2024.

\bibitem{goodfellow2014generative}
Ian~J Goodfellow, Jean Pouget-Abadie, Mehdi Mirza, Bing Xu, David Warde-Farley, Sherjil Ozair, Aaron Courville, and Yoshua Bengio.
\newblock Generative adversarial nets.
\newblock {\em Advances in neural information processing systems}, 27, 2014.

\bibitem{vaswani2017attention}
Ashish Vaswani, Noam Shazeer, Niki Parmar, Jakob Uszkoreit, Llion Jones, Aidan~N Gomez, {\L}ukasz Kaiser, and Illia Polosukhin.
\newblock Attention is all you need.
\newblock {\em Advances in neural information processing systems}, 30, 2017.

\bibitem{krizhevsky2009learning}
Alex Krizhevsky, Geoffrey Hinton, et~al.
\newblock Learning multiple layers of features from tiny images.
\newblock 2009.

\bibitem{deng2009imagenet}
Jia Deng, Wei Dong, Richard Socher, Li-Jia Li, Kai Li, and Li~Fei-Fei.
\newblock Imagenet: A large-scale hierarchical image database.
\newblock In {\em 2009 IEEE conference on computer vision and pattern recognition}, pages 248--255. Ieee, 2009.

\bibitem{lyons1998japanese}
Michael Lyons, Miyuki Kamachi, and Jiro Gyoba.
\newblock The japanese female facial expression (jaffe) dataset.
\newblock {\em (No Title)}, 1998.

\bibitem{lecun1998gradient}
Yann LeCun, L{\'e}on Bottou, Yoshua Bengio, and Patrick Haffner.
\newblock Gradient-based learning applied to document recognition.
\newblock {\em Proceedings of the IEEE}, 86(11):2278--2324, 1998.

\bibitem{he2016deep}
Kaiming He, Xiangyu Zhang, Shaoqing Ren, and Jian Sun.
\newblock Deep residual learning for image recognition.
\newblock In {\em Proceedings of the IEEE conference on computer vision and pattern recognition}, pages 770--778, 2016.

\bibitem{dosovitskiy2020image}
Alexey Dosovitskiy, Lucas Beyer, Alexander Kolesnikov, Dirk Weissenborn, Xiaohua Zhai, Thomas Unterthiner, Mostafa Dehghani, Matthias Minderer, Georg Heigold, Sylvain Gelly, et~al.
\newblock An image is worth 16x16 words: Transformers for image recognition at scale.
\newblock {\em arXiv preprint arXiv:2010.11929}, 2020.

\end{thebibliography}
\bibliographystyle{unsrt}

\newpage
\appendix
\section{Notation}

\bgroup
\def\arraystretch{1.5}

\begin{tabular}{p{1in}p{4in}}
$\displaystyle \gL$ & Loss objective function of the leaked model\\
$\displaystyle \gL^{(gen)}$ & Loss objective function of GIT\\
$\displaystyle  \sigma(\cdot) $ & The elementwise activation function\\
$\displaystyle  \sigma'(\cdot) $ & The derivative of $\sigma(\cdot)$\\
$\displaystyle  \vz $ & Pre-activation of an MIMO layer\\
$\displaystyle  \va $ & Post-activation of an MIMO layer\\
$\displaystyle \rmW$ & Weight matrix in a neural network\\
$\displaystyle \vg$ & Gradient of the loss objective function w.r.t $\rmW$\\
$\displaystyle B$ & Batch size in mini-batch training\\
$\displaystyle b$ & Batch index in mini-batch training\\
$\displaystyle \E$ & Empirical average over the samples in batch $b$\\
$\displaystyle d$ & Number of hidden nodes of an MIMO layer\\
$\displaystyle N_{in}$ & Number of input layers of an MIMO layer\\
$\displaystyle N_{out}$ & Number of output layers of an MIMO layer\\
$\displaystyle N$ & Number of input-gradient pairs for training GIT\\
$\displaystyle N'$ & Number of input-gradient pairs for testing\\
$\displaystyle  (\cdot)^+ $ & Moore-Penrose inverse of a matrix; or Moore-Penrose of each of $(\cdot)$'s subspace via the first dimension when $(\cdot)$ is a third order tensor\\
$\displaystyle  (\cdot)^T $ & Transpose of a matrix; or transpose of the second and the third dimension when $(\cdot)$ is a third order tensor\\
$\displaystyle  \otimes $ & Tensor Multiplification\\
$\displaystyle  \odot $ & Broadcast row-wise product\\
$\displaystyle  f(\cdot)^{(in)}$ & A module that approximates the input mapping of an MIMO layer\\
$\displaystyle  f(\cdot)^{(out)}$ & A module that approximates the output mapping of an MIMO layer\\
$\displaystyle  f'(\cdot)$ & The derivative of module $f(\cdot)$\\
$\displaystyle  \gD $ & Input-gradient pairs\\
$\displaystyle  E $ & Epoch budget\\
$\displaystyle  \eta $ & Learning rate\\
$\displaystyle  (\cdot)^{(i)} $ & The $i$-th sample in the dataset\\
$\displaystyle  M $ & The generative model GIT\\
$\displaystyle  \Theta $ & Trainable parameters of GIT\\
$\displaystyle  m_{\theta} $ & A shallow MLP parameterized by $\theta$ to approximate recursive reconstruction in Coarse-GIT\\
$\displaystyle  \vartheta $ & Trainable parameters in Module-GIT\\
$\displaystyle  \vx $ & Input data of the leaked model\\
$\displaystyle  \hat{\vx} $ & The estimated input data by GIT\\
\end{tabular}

\newpage
\section{Methodology Details}

\subsection{Exact-GIT}
\label{app:exact_git}

\textbf{Activation Function.} The Exact-GIT method in Algorithm~\ref{alg:general_train} requires iteratively applying Equation~(\ref{eq:reconstruct}). Equation~(\ref{eq:reconstruct}) involves the derivative of the activation function $\sigma'(\vz)$, which can be estimated by $\va$.
Although function $\sigma$ may not be an injective function, we demonstrate in Table~\ref{app_tab:func} below that we can uniquely identify $\sigma'(\vz)$ given $\va$ for the most popular activation functions used in practice.
\begin{table}[!ht] 
\centering
\begin{tabular}{p{1.8cm}<{\centering}|p{2.2cm}<{\centering}p{2.2cm}<{\centering}p{1.8cm}<{\centering}p{1.8cm}<{\centering}}
\Xhline{4\arrayrulewidth}
Name & ReLU & Leaky ReLU & Sigmoid & Tanh \\
\hdashline
$\va = \sigma(\vz)$ & $\max(0, \vz)$ & $\max(k\vz, \vz)$ & $\frac{1}{1 + e^{-\vz}}$ & $\frac{e^{\vz} - e^{-\vz}}{e^{\vz_i} + e^{-\vz}}$ \\[2pt]
$\sigma'(\vz)$ & $\begin{cases}
    1 & \text{if } \va > 0 \\
    0 & \text{if } \va = 0 \\
\end{cases}$ & $\begin{cases}
    1 & \text{if } \va > 0 \\
    k & \text{if } \va \leq 0 \\
\end{cases}$ & $\va_i (1 - \va)$ & $1 - \va^2$\\[2pt]
\Xhline{4\arrayrulewidth}
\end{tabular}
\caption{Mappings from $\va$ to $\sigma_i'(\vz)$ for popular activation functions. Operations are elementwise.}
\label{app_tab:func}
\end{table}
In Exact-GIT, the weights of the generative attack model represent the estimated weights of the leaked model. Therefore, we can compare the difference between their weights to investigate to which degree the generative attack models recover the gradient-to-input inversion.
In this context, we run Exact-GIT based on Algorithm~\ref{alg:general_train} and plot its convergence curve as in Figure~\ref{app_fig:exact_git}. Figure~\ref{app_fig:exact_git} illustrates the $l_2$ distance curve between the generative model's weights and the leaked model's weights, alongside the MSE between the reconstructed inputs and the ground truth inputs.
As shown in Figure~\ref{app_fig:exact_git}, when Exact-GIT converges, its weights align closely with the ground truth weights. This convergence highlights the effectiveness of exact-GIT in extracting weight information from the leaked model.

\begin{figure}[H]
    \centering		 
    \includegraphics[width=0.8\linewidth]{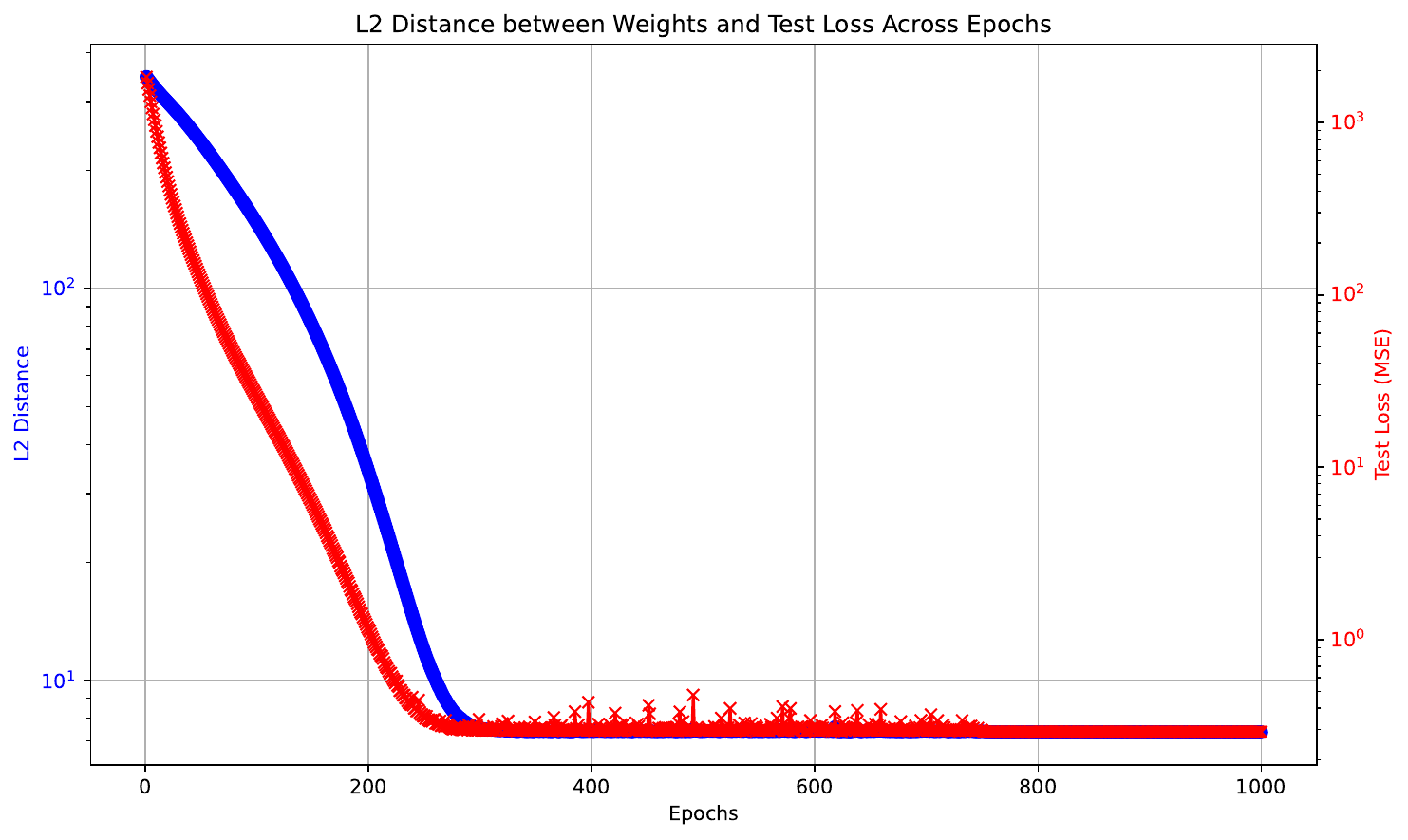}
    \vspace{-0.3cm}
    \caption{The red curve represents convergence curve of $l_2$ distance between weights of the generative model and the leaked model. The blue curve represents the convergence curve of MSE between reconstructed input and the ground truth input. The experiment is conducted on CIFAR-10 using Exact-GIT.}
    \label{app_fig:exact_git}
\end{figure}

\subsection{Coarse-GIT for Different Architectures}
\label{app:architecture}

Section~\ref{sec:analysis} demonstrates a generic framework for any neural network architectures as long as they support back-propagation.
In this section, we provide more technical details for specific neural network architecture that we use in the experiments, including feed forward networks, residual networks (ResNet) and vision transformer (ViT).
We believe the details in this section will provide more insights for practitioners to understand how GIT is adapted to different neural architectures.
Due to the scale of the architectures discussed in this section, we employ Coarse-GIT for all of them.

While related to the notation we use in Section~\ref{sec:analysis}, we use specific notations in this section for better readability. We provide the exact definition for each of these notations.

\subsubsection{Feed Forward Neural Networks} 
\label{app:feed_forward}

Feed forward neural networks, including multi-layer perceptron (MLP) and convolutional neural networks (CNN), can be formulated as follows:
\begin{equation} \label{app_eq:mlp&cnn}
\begin{aligned}
    &\gL_\theta(\vx, y) = \ell(\vz_N, y ) = \ell(\rmW_N \va_{N - 1}, y) \\
    &\va_i = \sigma_i(\vz_i),\ \vz_i = \rmW_i \va_{i - 1},\ i = 1, 2, ..., N - 1 \\
\end{aligned}
\end{equation}
We denote the number of hidden nodes for the $i$-th layer as $\{d_i\}_{i = 1}^{N - 1}$. 
The input data batch $\va_0 = \vx \in \mathbb{R}^{B \times d_0}$, where $B$ is the batch size.
$\theta = \{\rmW_i \in \mathbb{R}^{d_i \times d_{i-1}}\}_{i = 1}^N $ refer to the parameters of $N$ linear layers, including convolutional layers and fully connected layers. $\{\sigma_i\}_{i = 1}^{N - 1}$ are the nonlinear activation functions of different layers. In this context, $\{\vz_i \in \mathbb{R}^{B \times d_i}\}_{i = 1}^{N - 1} \text{and} \{\va_i \in \mathbb{R}^{B \times d_i}\}_{i = 1}^{N - 1}$ represent the pre-activation and post-activation of intermediate layers, respectively. $\vz_N = \rmW_N \va_{N -1}$ is the output logit, and $\ell$ is the function calculating the classification error, such as the softmax cross-entropy function.
We use $\vg_i = \nabla_{\rmW_i} \gL_\theta(\vx, y)$ to represent the gradient of each weight matrix. 

In this context, similar to Equation~(\ref{eq:grad}) and Equation~(\ref{eq:reconstruct}) in Section~\ref{sec:analysis}, we derive the back-propagation and then the iterative input layer approximation for feed forward neural network defined in Equation~(\ref{app_eq:mlp&cnn}) as follows:
\begin{equation} \label{app_eq:grad}
\begin{aligned}
    \vg_i =  \prod_{j = i}^{N - 1}  \left(\rmW_{j + 1}^T \odot \sigma_j'(\vz_j)\right) \otimes \frac{\partial \gL}{\partial \vz_N} \otimes \va_{i - 1}^T
\end{aligned}
\end{equation}
\begin{equation} \label{app_eq:reconstruct}
\begin{aligned}
    \va^T_{i - 1} \simeq \va_i^T \otimes \vg_{i + 1}^+ \otimes (\rmW_{i + 1}^T \odot \sigma_i'(\vz_i))^+ \otimes \vg_i  
\end{aligned}
\end{equation}
$\otimes$ and $\odot$ have the same definition as in Section~\ref{sec:analysis}.
As we can see, Equation~(\ref{app_eq:grad}) can be considered as a specific case of Equation~(\ref{eq:reconstruct}) where $N_{in} = N_{out} = 1$
Furthermore, we employ Coarse-GIT in the experiments. Specifically, we use an MLP model $f$ parameterized by $\vartheta$ to estimate $\va_{i - 1}$ from $\va_i$, $\vg_{i + 1}$ and $\vg_i$. We apply Equation~(\ref{app_eq:grad}) recursively and utilize it to reconstruct the input data.
\begin{equation} \label{app_eq:coarse_grad}
    \va_{i - 1} = f_\vartheta (\va_i, \vg_{i + 1}, \vg_i)
\end{equation}

\subsubsection{Residual Networks}

The key feature for residual networks (ResNet)~\cite{he2016deep} is the skip connections, resulting in $N_{out} > 1$ for layers that combine inputs from both the previous layer and shortcut connections.

Without the loss of generality, we generally follow the notation of feed forward neural network defined in (\ref{app_eq:mlp&cnn}) except that there is a single shortcut connection linking the $k$-th layer to $l$-th layer ($k < l$).
Specifically, the shortcut connection links the post-activation $\va_k$ to the pre-activation $\vz_l$ with a weight parameter $\rmS \in \sR^{d_k \times d_l}$.
Therefore, $\{\vz_i\}_{i = 1}^N$ and $\{\va_i\}_{i = 1}^N$ are calculated in the same manner except that $\vz_l = \rmW_l\va_{l - 1} + \rmS \va_{k}$.
Based on the back propagation, $\vg_i$ is calculated in the same way as in Equation~(\ref{app_eq:grad}) when $i > k$.
When $i \leq k$, $\vg_i$ is calculated as follows:
\begin{small}
\begin{equation} \label{eq:resnet}
\begin{aligned}
    \vg_i = &\prod_{j = i}^{k - 1}  M_j \otimes \left( \prod_{j = k}^{l - 1}  M_j + \rmS \odot \sigma'_k(\vz_k) \right) \otimes \prod_{j = l}^{N - 1}  \left( \rmW_{j + 1}^T \odot \sigma_j'(\vz_j) \right) \otimes
    \frac{\partial \gL}{\partial \vz_N} \otimes  \va_{i - 1}^T \\
\end{aligned}
\end{equation}
\end{small}
Following a similar analysis to feed forward neural networks, we can derive an approximation of $\va_{i - 1}$ using $\va_i$.
The approximation is the same as (\ref{app_eq:reconstruct}) except for the case $i = k$.
This is because we calculate $\va_i$ using $\va_{i - 1}$ in the same manner except for the case $i = k$, where the $k$-th layer is connected to not only the immediate preceding layer but also the $l$-th layer via shortcut connection.
Therefore, $\va_{k - 1}$ is approximated in a different way from~(\ref{app_eq:reconstruct}) as follows.
\begin{equation}
\begin{aligned} \label{app_eq:reconstruct_resnet}
     \va_{k - 1} \simeq \left( \rmW_{k +1}^T \odot \sigma'_k(\vz_k)\right) \otimes \vg_{k + 1} \otimes (\va_k^T)^+ + (\rmS \odot \sigma'_k(\vz_k)) \otimes \vg_l \otimes (\va_{l - 1})^+ \otimes \vg_k \\
\end{aligned}
\end{equation}

Compared with (\ref{app_eq:reconstruct}), the estimation in (\ref{app_eq:reconstruct_resnet}) incorporates not only $\vg_k$ and $\vg_{k + 1}$ but also $\vg_l$ to estimate $\va_{k - 1}^T$, which is consistent with the case of $N_{out} = 2$ in the analysis in Section~\ref{sec:analysis}. Since $\va_k$ is connected to $\vz_l$ via skip connection, gradients can flow directly from the $k$-th layer to the $l$-th layer in back propagation.
The insight provided by the approximation in Equation~(\ref{app_eq:reconstruct_resnet}) indicates that the reconstruction sequence follows the same path as the gradient flow during backpropagation.

We use Coarse-GIT in the experiment for ResNet, similar to Equation~(\ref{app_eq:coarse_grad}), we employ an MLP model $f$ parameterized by $\vartheta$ and reconstruct $\va_{k - 1}$ by:
\begin{equation}\label{app_eq:coarse_grad_resnet}
    \va_{k - 1} = f_\vartheta (\va_k, \vg_{k + 1}, \vg_{k}, \vg_l)
\end{equation}
When estimating the input from the leaked gradients, we apply Equation~(\ref{app_eq:coarse_grad_resnet}) when there is a shortcut connection and Equation~(\ref{app_eq:coarse_grad}) otherwise.

\subsubsection{Vision Transformer (ViT)}

In the case of vision transformer (ViT)~\citep{dosovitskiy2020image}, we apply modularized input data reconstruction and represent the each self-attention module as follows:
\begin{equation} 
\begin{aligned} \label{eq:attention}
    \vz = \text{softmax}\left(\frac{\rmQ \rmK^\top}{\sqrt{d_k}}\right) \rmV, \quad \rmQ = \va_i^{(in)} \rmW^Q,\quad \rmK = \va_i^{(in)} \rmW^K,\quad \rmV = \va_i^{(in)} \rmW^V
\end{aligned}
\end{equation}
where $\rmW^Q$, $\rmW^K$ and $\rmW^V$ represent the mapping weights to the tuple of query, key and value.
In multi-head attention (MHA), we concatenate the outputs of several self-attention modules and transform them by an affine operation. Without the loss of generality, we focus on single layer attention.
Furthermore, we reorganize Equation~(\ref{eq:attention}) to fit the formulation of Equation~(\ref{eq:module}):
\begin{equation}
    \begin{aligned}
        \vz = f_i^{(in)}([\rmQ, \rmK, \rmV]) := \text{softmax}\left(\frac{\rmQ \rmK^\top}{\sqrt{d_k}}\right) \rmV \\ [\rmQ, \rmK, \rmV] = \va^{(in)}_i \rmW^{(in)}_i := \va^{(in)}_i [\rmW^Q, \rmW^K, \rmW^V]
    \end{aligned}
\end{equation}
Equation~(\ref{eq:attention}) identifies the concrete definitions of $f^{(in)}$ and $\rmW^{(in)}_i$ for self-attention modules in the framework by Equation~(\ref{eq:module}) so that we can plug these definitions employ Equation~(\ref{eq:reconstruct_module}) to reconstruct the input of the attention layer by the leaked gradients.

Due to the large amount of parameters in ViT, we use Coarse-GIT for input reconstruction. If we use $\vg^Q$, $\vg^K$, $\vg^V$ to represent the leaked gradients of $\rmW^Q$, $\rmW^K$ and $\rmW^V$, respectively, then we employ an MLP module $f$ parameterized by $\vartheta$ to reconstruct $\va_i^{(in)}$ in a self-attention module.
\begin{equation}
    \va^{(in)}_i = f_\vartheta (\vg^Q, \vg^K, \vg^V, \{\vg_j^{(out)}\}_{j = 1}^{N_{out}}, \vz)
\end{equation}
where $\{\vg_j^{(out)}\}_{j = 1}^{N_{out}}$ are the leaked gradients of output matrices as defined for a general MIMO layer in Section~\ref{sec:analysis}.
The gradient inversion of fully-connected layers and residual structure in the ViT follows the same formulation as described in previous sections. Altogether, we can iteratively employ these formulas to reconstruct the input estimation of each layer, starting from the last layer and progressing to the first layer of the ViT model, eventually obtaining the input data estimation.


\section{Experiment Configurations} \label{app:experiment_setting}

\textbf{Universal Settings} We employ various architectures for the leaked model. For LeNet, we use a 5-layer configuration with kernel size 2 and same padding. For ResNet, we adopt a 15-layer variant with kernel size 3, consisting of 4 blocks, each containing 2 convolutional layers and 1 skip connection. For ViT, we connect four 4-head attention blocks following the patch embedding layer.

For generative methods, we use $10000$ batches of input-gradient pairs from the public dataset to train the generative model. During reconstruction, we use $10000$ batches of gradients from the local dataset to recover the corresponding local data.
For iterative optimization-based methods, we perform reconstruction for each batch of local data by starting from dummy inputs and applying iterative optimization individually.

For Coarse-GIT and Module-GIT, we use $m_\theta$ and $f_\vartheta$ with $3000$ neurons in each hidden layer. For LTI, we employ a generative model consisting of three hidden layers, each with $3000$ neurons, as described in~\cite{wu2023learning}.

\textbf{Specific Settings} In our experiments described in Section~\ref{subsec:direct}, the inference time for generative methods is computed as the average over reconstructing $10000$ local data batches, whereas for optimization-based methods, it is calculated based on the average over $10$ local data batches, since each reconstruction is significantly more time-consuming.

In our experiments described in Section~\ref{subsec:hybrid}, the inference time of generative+optimization-based hybrid methods is computed as the sum of their individual inference times. For all methods, inference time is calculated as the average over $10$ local data batches.

\section{More Experimental Analyses and Ablation Studies}
\label{app:anlysis}

\subsection{Reconstruction with Clipped Gradients}
\label{app:prune}

The prune rate $\gamma$ represents the proportion of gradient directions with small absolute values that are pruned (Pruning is applied by a mask with 0 and 1 values, therefore the dimension of gradients is not changed). As shown in Table~\ref{table:prune}, gradient pruning has minimal impact on GIT's performance but significantly degrades the performance of DLG. The results indicate that even when pruning 90\% of the gradient components with smaller absolute values, generative approaches remain largely unaffected, relying only on the top 10\% of the largest gradient values for training. This suggests that generative approaches primarily capture the dominant gradient components with large absolute values during training, unlike optimization-based methods, which require a finer alignment with the full gradient information.

It can be deduced from Table~\ref{table:prune} that generative approaches are less effective than optimization-based methods in recovering fine-grained image details. However, they demonstrate greater robustness against inaccurate gradients and gradient pruning while also being more efficient. Furthermore, generative methods train significantly faster, as gradient matching requires extensive computation to precisely align finer gradient details, leading to higher time complexity.

\begin{table}[H]
\centering
\caption{Comparison of the MSE under gradient pruning with varying pune rate. The dataset is CIFAR10 and the leaked model is LeNet.} 
\begin{tabular}{c|cccc}
    \Xhline{4\arrayrulewidth}
    \textbf{\textbf{Prune rate $\gamma$}} & \textbf{DLG} & \textbf{LTI} & \textbf{GIAS} & \textbf{GIT}\\
    \hline
    0 & 0.073 & 0.015 & 0.027 & \textbf{0.010} \\
    0.9 & 0.098 & 0.016 & 0.033 & \textbf{0.010} \\
    0.99 & 0.116 & 0.021 & 0.049 & \textbf{0.016} \\
    0.999 & 0.187 & 0.049 & 0.070 & \textbf{0.040} \\
    \Xhline{4\arrayrulewidth}
\end{tabular}
\label{table:prune}
\end{table}

\subsection{Effect of Noise on the Performance of Optimization-Based Reconstruction Methods}

Under inaccurate gradients, generative approaches demonstrate robust performance, as shown in Table~\ref{table:noise}. However, IG fails to recover meaningful data with as little as $0.01$ noise applied. This highlights the significant impact of noise on gradient matching methods like DLG.
In the contrast, IG fails to recover meaningful data with as little as $0.01$ noise applied. This highlights the significant impact of noise on optimization-based methods like IG.

\begin{figure}[H]
    \centering		 
    \includegraphics[width=1.0\linewidth]{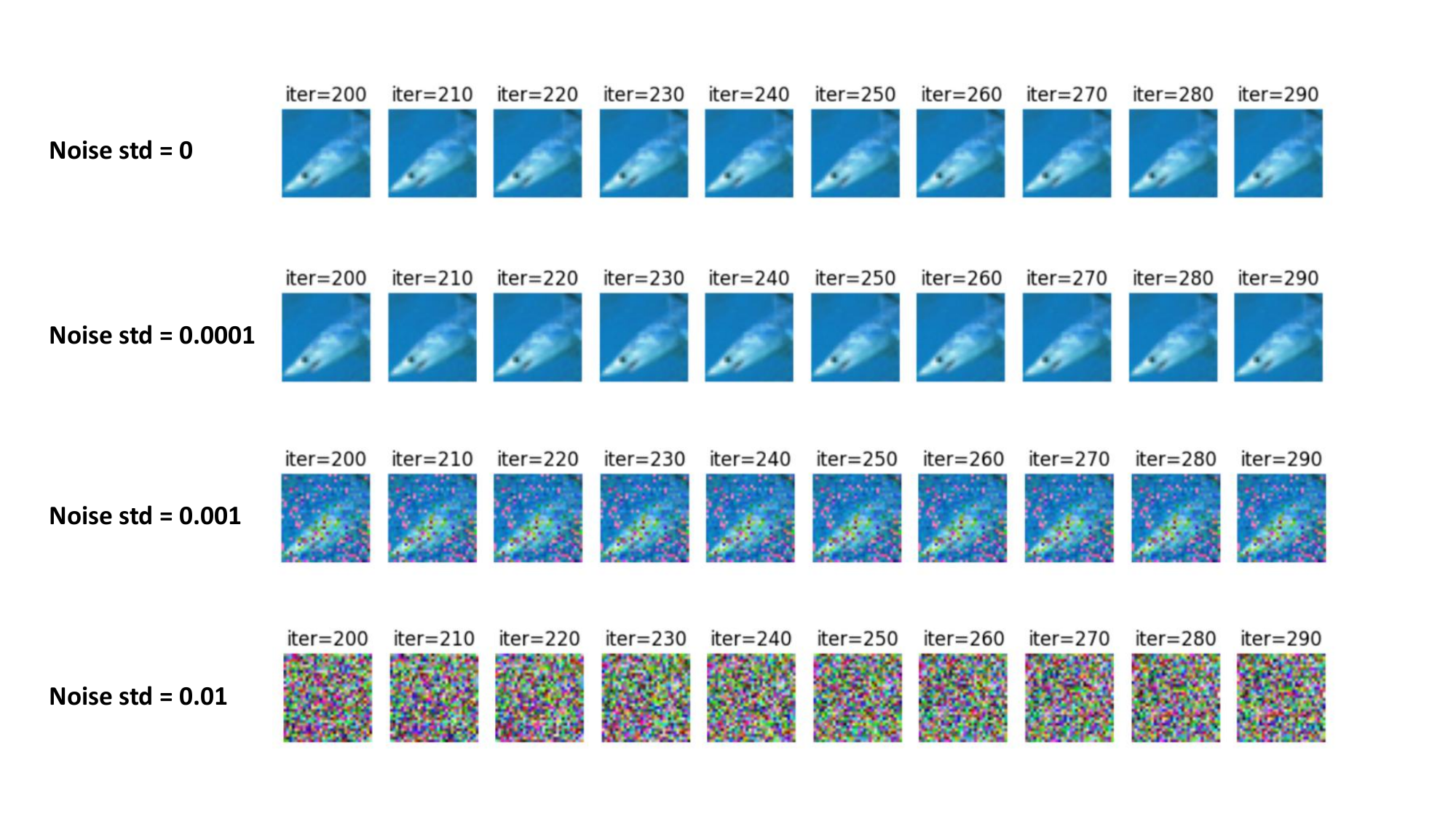}
    \caption{The figure illustrates the reconstructed images for IG when the leaked model is LeNet and the dataset is CIFAR-10. Varying levels of noise are applied to the gradients. The results depict IG’s reconstructions between the 200th and 300th optimization iterations.}
    \label{fig:noise}
\end{figure}

\subsection{Reconstruction without gradient of last layer's bias}

When the last layer of the neural network has a bias term $\vb_N$, i.e., $\va_N = \rmW_N \va_{N - 1} + \vb_N$, following the idea of~\cite{ma2023instance}, we have $\frac{\partial \gL}{\partial \vz_N} = \frac{\partial \gL}{\partial \vb_N}$. That is to say, we can directly utilize the gradient of the bias term in the last year as $\frac{\partial \gL}{\partial \vz_N}$.
When the last layer of the neural network does not have a bias term, we cannot directly obtain $\frac{\partial \gL}{\partial \vz_N}$. Therefore, we employ ablation study using the leaked gradients for the weight of the last layers to estimate the output logits by a shallow MLP.

\begin{table}[!ht]
    \centering
    \caption{Quantitative comparison for GIT with and w/o gradient of last layer's bias.}
    \renewcommand{\arraystretch}{1.2}
    \begin{tabular}{l!{\vrule width 1.1pt}c!{\vrule width 1.1pt}c!{\vrule width 1.1pt}cccc}
        \Xhline{4\arrayrulewidth}
        Dataset & \begin{tabular}{l} Leaked \\ Model \end{tabular} & Method & MSE$\downarrow$ & PSNR$\uparrow$ & LPIPS$\downarrow$ & SSIM$\uparrow$ \\ \Xhline{4\arrayrulewidth}
        \multirow{2}{*}{CIFAR10} 
        & \multirow{2}{*}{LeNet}  & with bias   & 0.010 & 20.38 & 0.2663 & 0.5533 \\
        \cdashline{3-7}
        &                         & w/o bias    & 0.012 & 19.35 & 0.2879 & 0.5035 \\
        \Xcline{1-7}{0.9pt}
        \multirow{2}{*}{Imagenet} 
        & \multirow{2}{*}{Resnet}  & with bias   & 0.039 & 14.42 & 0.8513 & 0.3507  \\
        \cdashline{3-7}
        &                         & w/o bias    & 0.039 & 14.30 & 0.9017 &  0.3120 \\
        \Xhline{3\arrayrulewidth}
    \end{tabular}
    \label{app_table:w/o_bias}
\end{table}

\subsection{Optimization-based methods with and w/o Generated Image Prior}
\label{app:curve}

As shown in the Figure~\ref{fig:hybrid}, the blue curve represents the convergence of the hybrid method, while the red curve illustrates IG without an image prior. It is evident that the hybrid method not only converges faster but also achieves superior performance. The fluctuations in the blue convergence curve are due to the small learning rate set for the optimizer, which causes oscillations when the loss falls below 1.

\begin{figure}[H]
    \centering		 
    \includegraphics[width=0.7\linewidth]{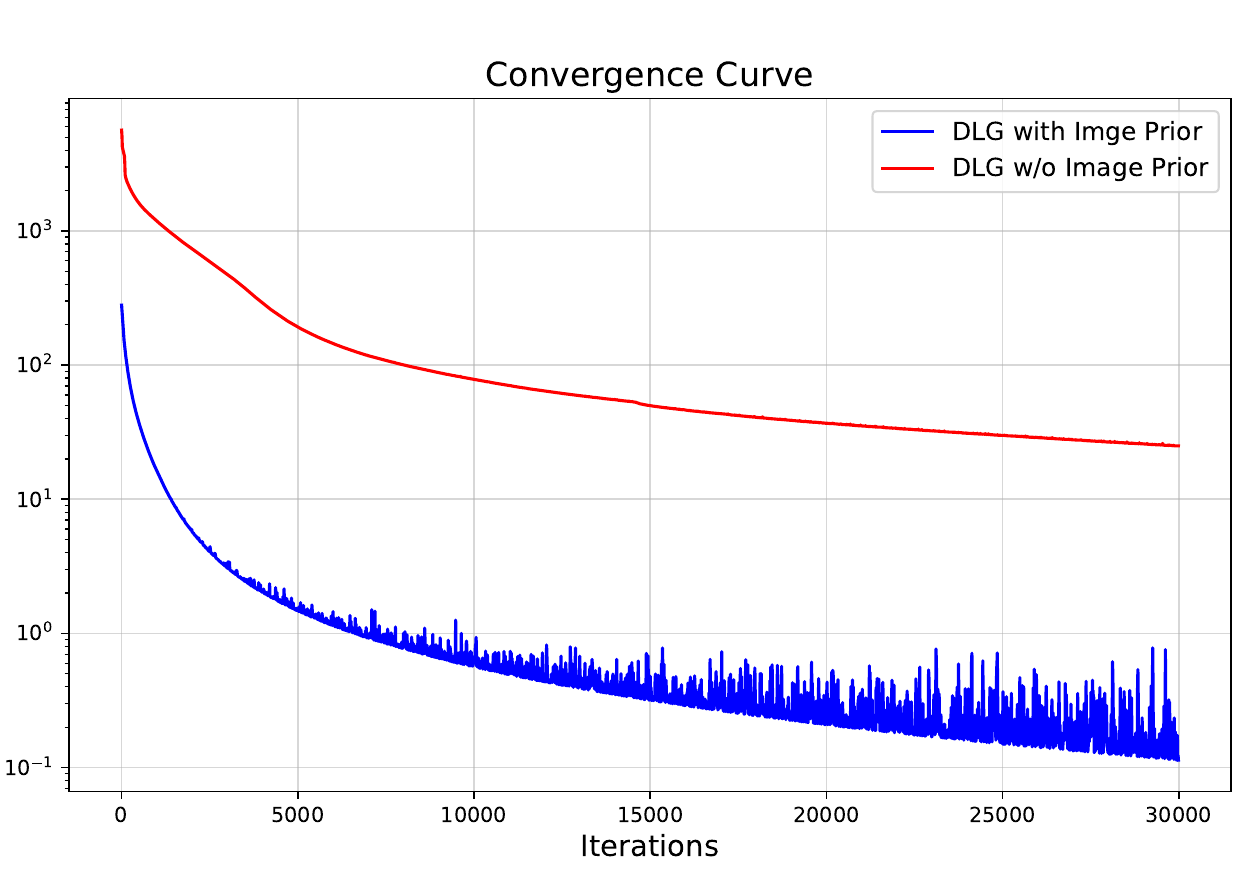}
    \vspace{-0.3cm}
    \caption{The convergence curve of DLG with and without an image prior. The leaked model is ResNet. The vertical axis indicates the distance between the dummy gradients and the corresponding ground-truth gradients.}
    \label{fig:hybrid}
\end{figure}

\subsection{Visual Results}
\label{app:visual} 

\subsubsection{Visual Results on CIFAR-10 and Tiny ImageNet}

Figure~\ref{app_fig:prior} illustrates reconstructed CIFAR-10 and Tiny ImageNet direct using GIT for reconstruction or using GIT as generated image prior. These results show that directly using GIT for reconstruction can achieve reasonable recovery but tends to lose some high-frequency details. In contrast, using GIT as an image prior—specifically for initializing optimization-based methods—helps preserve high-frequency information and achieves reconstruction quality beyond what optimization-based methods alone can attain.

\begin{figure}[H]
    \centering		 
    \includegraphics[width=1\linewidth]{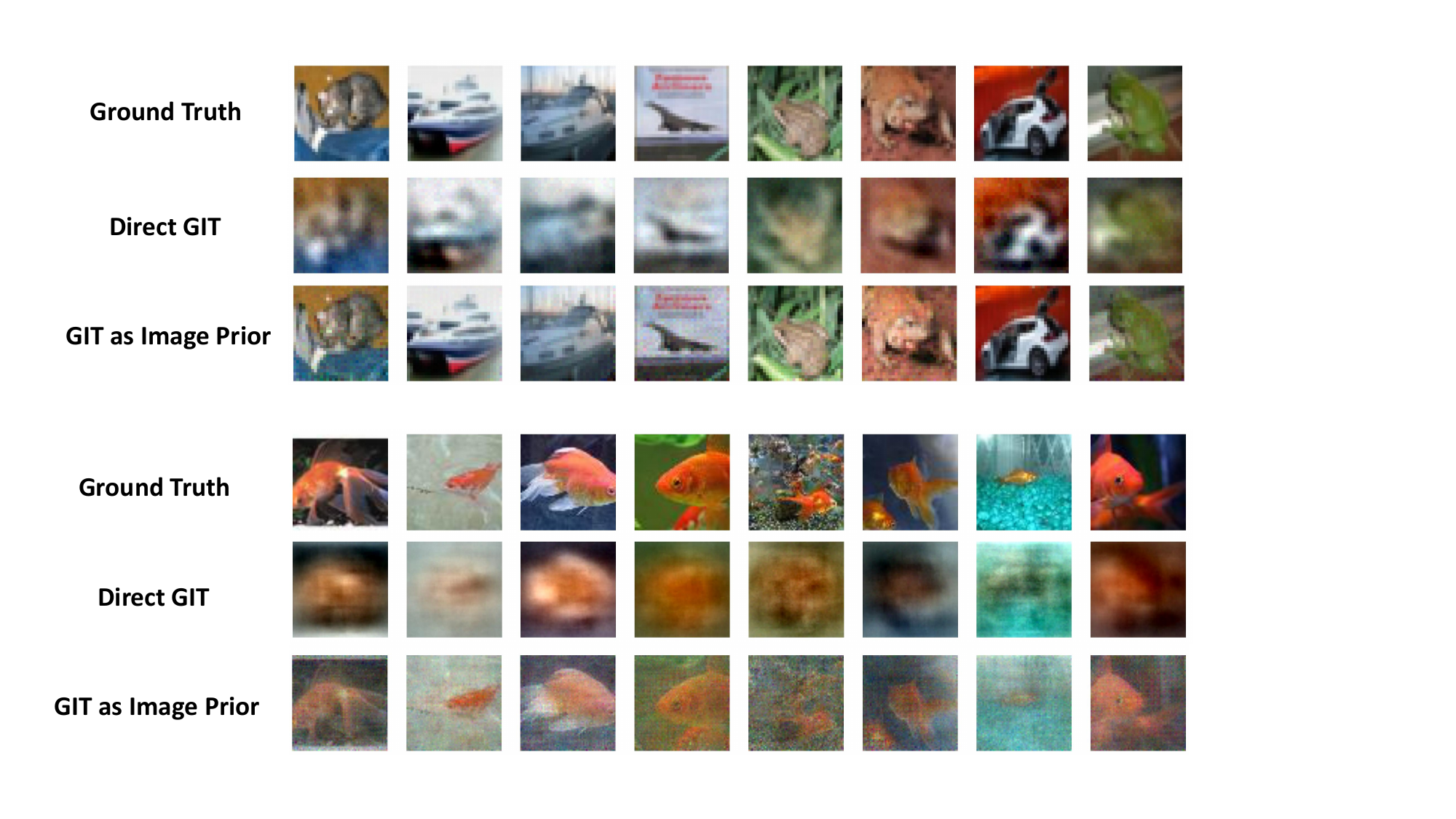}
    \caption{The figure illustrates the \textbf{ground truth input images}, the \textbf{direct reconstructed images by GIT} and the \textbf{reconstructed images using IG initialized with GIT-generated prior}, from top to the bottom respectively. The top three rows correspond to the CIFAR-10 dataset with the leaked model being LeNet, while the bottom three rows correspond to the TinyImageNet dataset with the leaked model being ResNet.}
    \label{app_fig:prior}
\end{figure}

\subsubsection{Visual Results for GIT on Large Resolution}

Reconstruction at high resolution tends to be more challenging, especially for images containing complex objects. To illustrate the characteristics of both easy and hard-to-recover examples, we present the first 8 and the best 100 reconstructions. Odd-numbered rows show the ground-truth images, while even-numbered rows display the corresponding reconstructions obtained directly using GIT.

\begin{figure}[ht]
    \centering		 
    \includegraphics[width=0.8\linewidth]{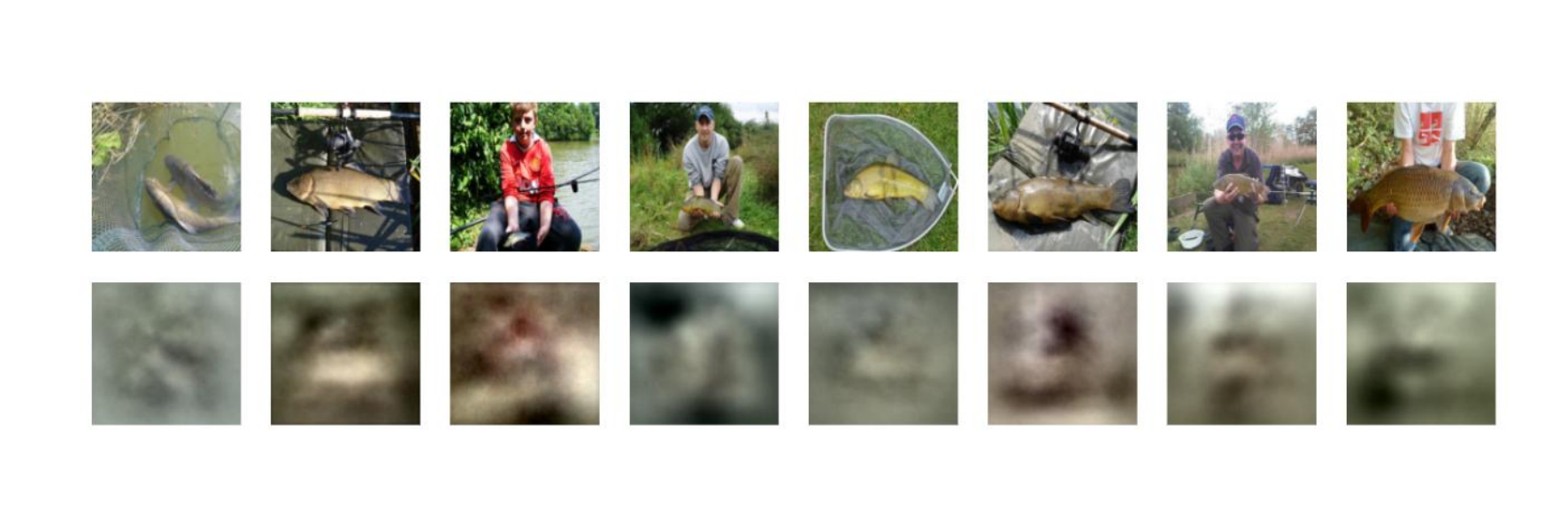}
    \vspace{-0.3cm}
    \caption{The first 8 reconstructed images (ImageNet, ResNet). Odd-numbered rows show the ground-truth images, while even-numbered rows display the corresponding reconstructions obtained directly using GIT.}
    \label{fig:imagenet_resnet_first8}
\end{figure}

\begin{figure}[ht]
    \centering		 
    \includegraphics[width=0.6\linewidth]{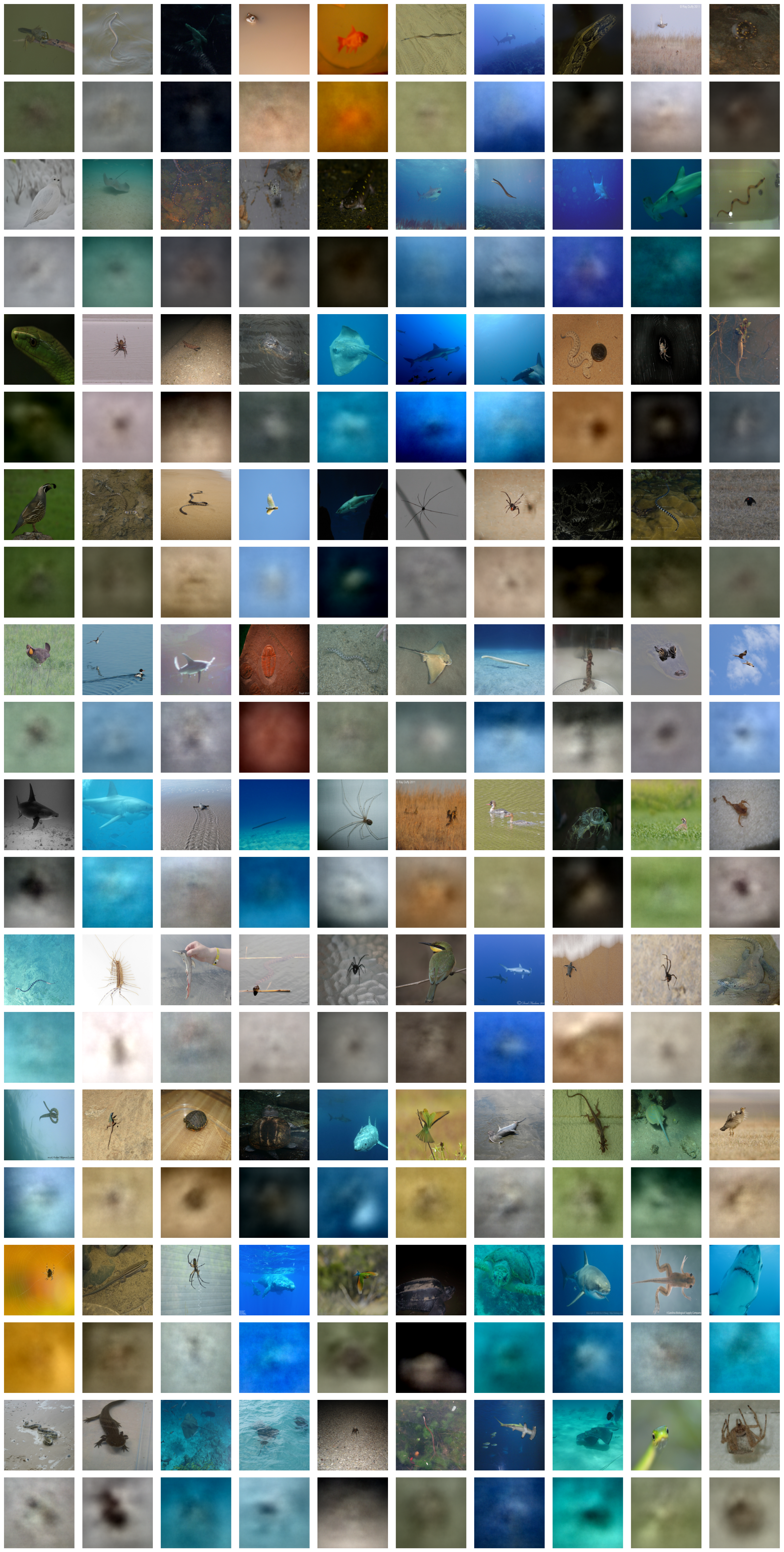}
    \vspace{-0.3cm}
    \caption{The best 100 reconstructed images with lowest MSE (ImageNet, ResNet). Odd-numbered rows show the ground-truth images, while even-numbered rows display the corresponding reconstructions obtained directly using GIT.}
    \label{fig:imagenet_resnet_best100}
\end{figure}

\end{document}